\documentclass[preprint,12pt]{elsarticle}
\usepackage{tikz}

\newcommand*\annotatedFigureText[4]{\node[draw=none, anchor=south west, text=#2, inner sep=0, text width=#3\linewidth,font=\sffamily] at (#1){#4};}
\newenvironment {annotatedFigure}[1]{\centering\begin{tikzpicture}
\node[anchor=south west,inner sep=0] (image) at (0,0) { #1};\begin{scope}[x={(image.south east)},y={(image.north west)}]}{\end{scope}\end{tikzpicture}}

\usepackage[numbers]{natbib}
\usepackage{graphicx} 
\usepackage{amsmath, amsfonts}
\usepackage{mathtools}
\usepackage{caption}
\usepackage{float}
\usepackage{setspace}
\usepackage{placeins}
\usepackage{amssymb}
\usepackage{caption}
\usepackage{hyperref}
\usepackage{booktabs}
\def\tsc#1{\csdef{#1}{\textsc{\lowercase{#1}}\xspace}}
\tsc{WGM}
\tsc{QE}

\begin{document}

\begin{frontmatter}

\title{Hybrid Latent-Structural Fusion (HLSF) for Cyber Anomaly Detection}  




\author[1]{Dorianis Mercedes Perez\corref{cor1}\fnref{fn1}}
\ead{dorpere@sandia.gov}
\fntext[fn1]{Present address: Center for Computing Research, Sandia National Laboratory, Albuquerque, NM 87185}
\affiliation[1]{organization={X Computational Physics Division},
            addressline={Los Alamos National Laboratory}, 
            city={Los Alamos},
            postcode={87545}, 
            state={NM},
            country={USA}}

\cortext[cor1]{Corresponding author}


\author[2]{Maksim Ekin Eren}
\ead{maksim@lanl.gov}
\affiliation[2]{organization={Analytics, Intelligence, and Technology Division},
            addressline={Los Alamos National Laboratory}, 
            city={Los Alamos},
            postcode={87545}, 
            state={NM},
            country={USA}}

\author[1]{Bryan Edward Kaiser}
\ead{bkaiser@lanl.gov}


\begin{abstract}
Malicious anomalous activity detection is a fundamental challenge for cyber security systems. Both tensor decomposition under statistical framework with CANDECOMP-PARAFAC alternating Poisson regression (CP-APR) and normalizing flows have proven to be powerful unsupervised machine learning methods that model multi-dimensional data and capture complex and multi-faceted details of behavior profiles in cyber security applications. In this study, we propose Hybrid Latent-Structural Fusion (HLSF), a weighted anomaly fusion framework integrating CP-APR structural anomaly scores with latent-space density scores derived from normalizing flows. In our experiments, we show that the HLSF framework improves anomaly detection performance on a dataset of real-world compromised user credentials collected from the large enterprise network of Los Alamos National Laboratory (LANL) during a red-teaming exercise, compared with using CP-APR or normalizing flows alone.

\end{abstract}



\end{frontmatter}

\section{Introduction}


Digital information has been vulnerable to attacks since the earliest days of computing. Networked systems first emerged in the 1960s and 1970s and so did early malware and cyber breaches \citep{DBLP:journals/corr/abs-2007-15759}. Anderson's report in 1980 introduced the idea that security audit trails can be examined for unauthorized access to systems \citep{anderson1980computer}. Detection of intrusions via anomaly-based approaches was then founded in 1987 when Denning formalized the idea of anomaly-based detection by monitoring system behavior against statistical profiles of normal usage \citep{denning1987intrusion}. Today, detection of cyber anomalies, such as compromised accounts, insider threats, malware traffic, and phishing, continues to be a significant challenge for cyber defenders \citep{eren2023general}. These challenges underscore the need for methods that can automatically identify subtle deviations from normal behavior in large, complex cyber environments. 

Recent work has demonstrated that latent-factor models based on dimensionality reduction, originally developed for recommendation systems with collaborative filtering, provide a powerful framework for modeling cyber behavior \citep{10.1145/138859.138867}. Matrix factorization can identify "peer groups" of users and devices, which allow data-driven predictions of future user actions \citep{ Conroy2018RedTeam, Turcotte2016PoissonFactorization, Passino2020GraphLinkPrediction}. More recently, tensor factorization methods have extended these ideas to higher-order dimensionality reduction on cyber data that capture interactions among users, devices, resources, and time simultaneously. Eren et al. showed that extending tensor factorization methods within a statistical framework enables state-of-the-art unsupervised anomaly detection across a range of cybersecurity applications \citep{eren2023general}. Their approach used an implementation of the CANDECOMP-PARAFAC
alternating Poisson regression (CP-APR) algorithm originally introduced by \cite{doi:10.1137/110859063} and achieved stronger anomaly detection performance than statistical matrix factorization, as well as other supervised and unsupervised methods, on tasks involving spam e-mail traffic, anomalous credit card transactions, botnet network traffic, and detecting users with compromised credentials.

In parallel, normalizing flows have emerged as an approach for out-of-distribution (OOD) detection in real-world applications spanning autonomous driving, medical diagnosis, and security systems \citep{ZHAO2026133081, Wang2025OODAutonomousDriving, Cao2020MedicalOODBenchmark, Hong2024MedicalOODSurvey, Chandola2009AnomalySurvey, Chalapathy2019DeepAnomalySurvey, Yang2024GeneralizedOODSurvey}. Normalizing flows have been recognized across the deep learning community to be a powerful tool because of their ability to model complex probability distributions and perform exact likelihood estimation \citep{ZHAO2026133081, Dinh2014NICE, Dinh2016RealNVP, Papamakarios2021NormalizingFlows, kobyzev2020normalizing}. Normalizing flows learn an exact, flexible density model via bijective mappings from data space to latent space \citep{Papamakarios2021NormalizingFlows, ZHAO2026133081}. This makes them effective in extracting features from in-distribution (ID) data by training them on high level features of ID data, via maximization of likelihoods, and then using low likelihood as a score function for OOD detection \citep{ZHAO2026133081, Nalisnick2019DoModelsKnow, Serra2020InputComplexity, Kirichenko2020WhyFlowsFail, Zhang2021UnderstandingOODFailures, Ahmadian2021LikelihoodFreeOOD}. Because of this, normalizing flows have become a popular and successful method in unsupervised anomaly detection across many applications \citep{zhou, HU2025111220, Gudovskiy_2022_WACV, Rudolph_2021_WACV}.

In this study, we propose the use of a hybrid approach, named Hybrid Latent-Structural Fusion (HLSF), where CP-APR allows the structured representation via latent factors of the tensorized sparse data and normalizing flows, via Real-valued Non-Volume Preserving (RealNVP), helps us model the density of the latent behavior \citep{Dinh2016RealNVP}. This makes the normalizing flow the detector of latent combinations not seen during training, which is the cyber anomaly signal. CP-APR learns what patterns exist while normalizing flows learn how likely a specific pattern configuration is. Combining them improves anomaly detection by capturing both structure and distribution. Our results make a compelling case that this methodology can outperform either method on its own. Starting with sparse tensors constructed from the publicly available Unified Host and Network Dataset, which contains anomalies involving users with compromised credentials recorded on the large real-world Los Alamos National Laboratory (LANL) network during a cyber red-teaming event, we use the CP-APR algorithm to decompose a specified-rank tensor via Poisson tensor factorization (CP-APR) \citep{Turcotte2017UnifiedHA}. The resulting latent factors are then used to score anomalies using p-values from the fitted model. The second approach uses normalizing flows to learn a transformation of the data into a latent distribution and produces anomaly scores via the negative log likelihood (NLL). The hybrid approach scores the anomalies via a weighted sum. These approaches were then compared and we found that in certain matrix and tensor configurations, where we select different variables to represent the dimensions in the tensor, normalizing flows and, subsequently, the hybrid approach tend to outperform the CP-APR-based anomaly scores for all datasets. By modeling CP-APR latent factors instead of the raw tensor, the flow operates on a structured, continuous representation where simplicity reflects meaningful patterns rather than sparsity. This prevents degenerate or random inputs from receiving artificially high likelihood, mitigating the known bias of normalizing flows toward low-complexity data \citep{ZHAO2026133081, Ren2019LikelihoodRatiosOOD, Nalisnick2019DoModelsKnow, Serra2020InputComplexity, Osada_Takahashi_Nishide_2024}. In summary, our contributions are as follows:
\begin{enumerate}
    \item Leveraged normalizing flows to detect cyber anomalies in network-flow data, focusing on users with compromised credentials in a real-world network dataset.
    \item Compared the anomaly detection performance of normalizing flows with tensor decomposition using CP-APR.
    \item Introduced a hybrid framework, named HLSF, that fuses anomaly scores from tensor decomposition and normalizing flows to improve detection performance.
\end{enumerate}

The remainder of the paper is organized as follows. Section 2 describes related work in anomaly detection methods and the motivation for choosing these specific methods, Section 3 describes our methodology and presents the mathematical formulations for each approach, Section 4 presents the experimental setup and the results, and Section 5 concludes the paper with key findings.

\section{Related Work}

Recent advances in cyber anomaly detection have increasingly focused on latent-factor models that capture underlying behavioral structure in large-scale security data \citep{Momtazpour2015, Lee03042022, challu22a, wang2024}. Early work used Poisson matrix factorization (PMF) to model interactions among users, hosts, and resources while identifying statistically unlikely events through Poisson distribution-based anomaly scores \citep{Turcotte2016PoissonFactorization, PriceWilliams2018TimeOfDay, Passino2020GraphLinkPrediction}. These approaches demonstrated that recommendation-system methodologies can effectively discover peer groups and behavioral patterns in an unsupervised fashion, without need of labels, that are difficult to identify using rule-based or signature-based techniques.

\begin{figure}
    \centering
    \includegraphics[width=0.5\linewidth]{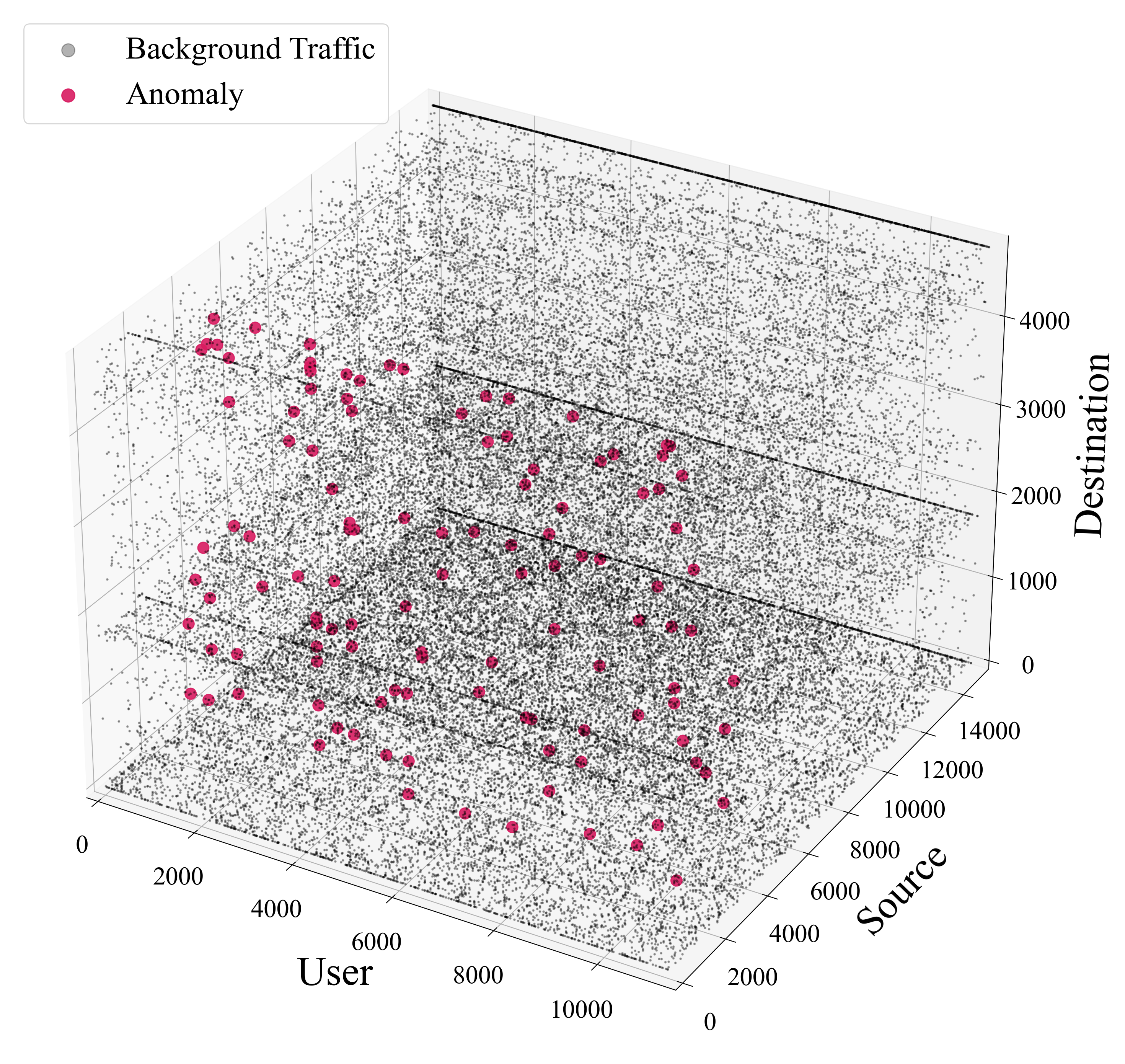}
    \caption{Binary tensor with User-Source-Destination dimensions showing background traffic in gray and anomalies in red from Eren et al. \citet{eren2023general} }
    \label{bin_tensor}
\end{figure}

Tensor factorization methods extended latent-factor modeling from pairwise relationships up to higher-order relationships. Cyber events are, by nature, multi-relational, involving users, devices, destinations, applications, and temporal attributes including cyclostationary processes simultaneously. Tensor representations model higher-order relationships with dimensionality reduction to uncover latent behavioral structure across dimensions. Eren et al. introduced a generalized version of a tensor decomposition method for unsupervised anomaly detection that surpasses state-of-the-art supervised and semi-supervised learning baselines across an array of challenging and diverse cyber application areas \citet{eren2023general}. They use binary tensors to represent user, source, and destination, for example Figure \ref{bin_tensor}, and perform tensor factorization via canonical polyadic decomposition (CPD) to obtain a weighted sum of rank-1 tensors. Their tensor analysis-based method is sensitive to anomalous activity over a diverse set of attributes. The publicly available \textit{pyCP\_APR} Python library they developed allows tensor factorization using a graphics processing unit (GPU) and combines factorization with an anomaly scoring and prediction interface \citep{eren2023general, Eren2020ISI, Eren2021pyCPAPR}. In this work, we use the Python library \textit{pyCP\_APR} for tensor factorization.

A key ability of CP-APR is modeling sparse count and binary cyber data through a Poisson likelihood formulation. Anomaly scores are derived from deviations relative to the reconstructed tensor representation which may not fully capture complex nonlinear dependencies among latent factors. For this reason, we explore complementary density-estimation approaches capable of learning richer probability distributions over latent behavioral representations.

Deep generative models have emerged as a powerful class of methods for unsupervised anomaly detection because they learn probability distributions directly from observed data. Variational autoencoders (VAEs), generative adversarial networks (GANs), and normalizing flows have all been applied to an array of cyber security problems, including intrusion detections, malware analysis, and network traffic monitoring \citep{Arora04032022, Shaikh2025, Taylor_Eleyan_Al-Khalidi_2025, durai, sayyad2024vae, nuiaa2025phishnetvae, ADIBAN2023296, dunmore, Stepashkina2024IndustrialCyberFlows, Ammar2025FlowAnomalyIDS, Guo2025FlowSemiSupervisedIDS}. Normalizing flows are particularly attractive because they provide exact likelihood estimation through an invertible mapping between observed data and a tractable latent distribution.

Recent studies have demonstrated the success of using normalizing flows for cyber anomaly detection. \citet{Stepashkina2024IndustrialCyberFlows} used a transformer-based autoencoder architecture enhanced by transitioning to a variational autoencoder (VAE) and integration of normalizing flows to detect anomalies in multivariate time series. \citet{Guo2025FlowSemiSupervisedIDS} used a semi-supervised learning framework with normalizing flows to model the distribution of both labeled and unlabeled data while incorporating sensitivity threshold management to enhance detection performance. \citet{Ammar2025FlowAnomalyIDS} used a RealNVP architecture trained using only benign traffic data to successfully detect anomalies including Denial of Service (DoS), Distributed Denial of Service (DDoS), port scans, infiltration, and botnet activities. 

While these studies demonstrate that normalizing flows can effectively model complex cyber data distributions, existing approaches primarily operate on network-flow features, packet statistics, or temporal sensor measurements. To the best of our knowledge, this is the first time flow-based density estimation are applied to latent behavioral representations derived from tensor factorization methods.

\section{Methods}
In this section, we describe the three methods compared in this study: CP-APR tensor factorization, RealNVP normalizing flows, and the proposed Hybrid Latent-Structural Fusion (HLSF) method. We first introduce the tensor notation used throughout the paper, since both CP-APR and HLSF operate on sparse, multi-way event tensors. We then describe CP-APR, the RealNVP density estimator, and finally the proposed fusion framework. Although CP-APR was first introduced by \cite{doi:10.1137/110859063} and later used for cyber anomaly detection by ~\cite{Eren2020ISI,eren2023general}, we include a brief description for completeness.

\subsection{Tensor Representation and Terminology}

The authentication data are represented as sparse count (pr binary) tensors. A tensor generalizes a matrix to more than two dimensions, or modes. Let
\begin{equation}
    \mathcal{X} \in \mathbb{R}^{N_1 \times N_2 \times \cdots \times N_D}
\end{equation}
denote a $D$-dimensional tensor, where $D$ is the number of modes and $N_d$ is the dimension size of mode $d$. The total number of possible tensor entries is
\begin{equation}
    |\mathcal{X}| = \prod_{d=1}^{D} N_d .
\end{equation}
Each entry is indexed by a multi-index
\begin{equation}
    \mathbf{i} = (i_1,\ldots,i_D),
\end{equation}
where $i_d \in \{1,\ldots,N_d\}$ identifies a specific entity in mode $d$. The corresponding tensor value is denoted by $\mathcal{X}_{\mathbf{i}}$ or equivalently $\mathcal{X}_{i_1,\ldots,i_D}$. For example, Figure~\ref{bin_tensor} shows a $D=3$ tensor with modes corresponding to \textit{User}, \textit{Source}, and \textit{Destination}. In this case, a tensor entry
\begin{equation}
    \mathcal{X}_{i_1,i_2,i_3}
\end{equation}
represents the observed count for user $i_1$ authenticating from source device $i_2$ to destination device $i_3$ within a specified time window. If the entry is nonzero, the corresponding event combination occurred at least once during that window. If the entry is zero, the combination was not observed.

Because cyber event tensors are extremely sparse, most possible combinations of users, devices, hosts, or other entities are never observed. We denote the number of nonzero entries in the tensor by
\begin{equation}
    nnz(\mathcal{X}) =
    \left|
    \left\{
    \mathbf{i} : \mathcal{X}_{\mathbf{i}} \neq 0
    \right\}
    \right| .
\end{equation}
The density of the tensor is therefore
\begin{equation}
    \rho(\mathcal{X})
    =
    \frac{nnz(\mathcal{X})}{\prod_{d=1}^{D} N_d},
\end{equation}
and the sparsity is
\begin{equation}
    1 - \rho(\mathcal{X})
    =
    1 -
    \frac{nnz(\mathcal{X})}{\prod_{d=1}^{D} N_d}.
\end{equation}
Thus, larger mode sizes increase the number of possible entries multiplicatively, while $nnz(\mathcal{X})$ counts only the entries that are actually observed. This distinction is important because the methods considered here are designed for sparse tensors where $nnz(\mathcal{X}) \ll \prod_{d=1}^{D} N_d$.

\subsection{CP-APR Tensor Factorization}

CP-APR is an efficient algorithm for optimizing sparse Poisson likelihood functions~\cite{doi:10.1137/110859063}. It is based on the Poisson canonical polyadic decomposition (CPD), which represents a sparse count tensor using a low-rank factorization. For a $D$-dimensional tensor with shape $N_1,\ldots,N_D$, each entry is modeled as an independent draw from a Poisson distribution:
\begin{equation}
    \mathcal{X}_{\mathbf{i}} \sim \operatorname{Poisson}(\lambda_{\mathbf{i}}),
\end{equation}
where the Poisson rate $\lambda_{\mathbf{i}}$ is determined by a rank-$R$ CPD:
\begin{equation}
    \lambda_{\mathbf{i}}
    =
    \sum_{r=1}^{R}
    \gamma_r
    \prod_{d=1}^{D}
    \theta^{(d)}_{r,i_d}.
\end{equation}
Here, $\gamma_r$ is the weight of component $r$, and $\theta^{(d)}_{r,i_d}$ is the value associated with entity $i_d$ in mode $d$ for component $r$. Equivalently, the fitted CP-APR model can be written as
\begin{equation}
    \Theta
    =
    \{A^{(1)},\ldots,A^{(D)},\gamma\},
\end{equation}
where each factor matrix
\begin{equation}
    A^{(d)}
    \in
    \mathbb{R}^{N_d \times R}
\end{equation}
contains the rank-$R$ latent factors for mode $d$.

The model is trained by maximizing the joint log-likelihood of the observed tensor counts, or equivalently by minimizing the generalized Kullback--Leibler divergence between the observed tensor and the fitted Poisson rates:
\begin{equation}
\log P(\mathcal{X})
=
\sum_{i_1=1}^{N_1}
\cdots
\sum_{i_D=1}^{N_D}
\left[
\mathcal{X}_{\mathbf{i}}
\log \lambda_{\mathbf{i}}
-
\log \Gamma(\mathcal{X}_{\mathbf{i}}+1)
\right]
-
\Lambda ,
\label{eq:cpapr_loglik}
\end{equation}
where
\begin{equation}
    \Lambda
    =
    \sum_{i_1=1}^{N_1}
    \cdots
    \sum_{i_D=1}^{N_D}
    \lambda_{\mathbf{i}} .
\end{equation}
Equation~\eqref{eq:cpapr_loglik} is computationally attractive for sparse tensors because the terms involving $\mathcal{X}_{\mathbf{i}}$ contribute only at nonzero entries, reducing the main likelihood computation to the observed entries~\citep{eren2023general}.

Rank selection is performed by fitting CP-APR models on the training set for ranks $R \in \{1,2,\ldots,100\}$ and evaluating the log-likelihood on the validation set. The rank with the highest validation log-likelihood is selected for subsequent training and testing. After rank selection, the training and validation sets are combined to fit the final CP-APR model~\citep{eren2023general}.

Because the tensors are extremely sparse, the fitted rank-$R$ model can assign zero or near-zero Poisson rates to some entries during testing. To reduce the effect of zero-rate estimates, the predicted Poisson rate for each entry is smoothed using a convex combination of the fitted rank-$R$ model and a rank-1 baseline model. Following Eren et al.~\citep{eren2023general}, binary entries are inflated so that the mean nonzero tensor value is approximately one:
\begin{equation}
    \mathcal{X}_{\mathbf{i}}
    =
    \frac{\prod_{d=1}^{D} N_d}{nnz(\mathcal{X})},
    \qquad
    \text{for } \mathcal{X}_{\mathbf{i}} \neq 0 .
\end{equation}
A rank-1 factorization is then estimated in addition to the selected rank-$R$ factorization. Since the marginal sums across tensor modes are nonzero in the training data, the rank-1 model provides nonzero baseline rates. The final smoothed Poisson rate is
\begin{equation}
    \lambda_{\mathbf{i}}
    =
    a \lambda^{1}_{\mathbf{i}}
    +
    (1-a)\lambda^{R}_{\mathbf{i}},
\end{equation}
where $\lambda^{1}_{\mathbf{i}}$ is the rank-1 rate, $\lambda^{R}_{\mathbf{i}}$ is the selected rank-$R$ rate, and $a \in [0,1]$ controls the amount of smoothing. Based on Eren et al.~\citep{eren2023general}, we set $a=0.1$.

CP-APR anomaly scores are computed using the Poisson survival function. For an observed test entry $\mathcal{X}_{\mathbf{i}}$, the p-value is the probability of observing a count at least as large as the observed value under the fitted model:
\begin{equation}
    P
    \left(
    X_{\mathbf{i}}
    \geq
    \mathcal{X}_{\mathbf{i}}
    \mid
    \lambda_{\mathbf{i}}
    \right).
\end{equation}
Lower p-values indicate events that are less likely under the learned low-rank relational structure. We define the CP-APR anomaly score as
\begin{equation}
s_{\mathrm{CP}}(\mathbf{i})
=
-\log
P
\left(
X_{\mathbf{i}}
\ge
\mathcal{X}_{\mathbf{i}}
\mid
\lambda_{\mathbf{i}}
\right).
\label{eq:cp_score}
\end{equation}

\subsection{Normalizing Flows}

Normalizing flows model a target distribution $p^{*}(\mathbf{x})$ using an invertible transformation of a tractable base distribution \cite{kobyzev2020normalizing, Papamakarios2021NormalizingFlows}. In this work, we use the RealNVP architecture. Let
\begin{equation}
    f_{\phi}: \mathbb{R}^{M} \rightarrow \mathbb{R}^{M}
\end{equation}
denote an invertible mapping from an input representation $\mathbf{x}$ to a latent variable
\begin{equation}
    \mathbf{z} = f_{\phi}(\mathbf{x}),
\end{equation}
where the latent variable follows a standard Gaussian base distribution,
\begin{equation}
    p_Z(\mathbf{z}) = \mathcal{N}(\mathbf{0},\mathbf{I}).
\end{equation}
By the change-of-variables formula, the likelihood of a sample is
\begin{equation}
\log p_X(\mathbf{x})
=
\log p_Z(f_{\phi}(\mathbf{x}))
+
\log
\left|
\det
\frac{\partial f_{\phi}}{\partial \mathbf{x}}
\right|.
\end{equation}

RealNVP constructs $f_{\phi}$ as a composition of $K$ affine coupling layers. For an input partitioned into $(\mathbf{x}_1,\mathbf{x}_2)$, one coupling layer is defined as
\begin{align}
\mathbf{y}_1 &= \mathbf{x}_1, \\
\mathbf{y}_2 &=
\mathbf{x}_2
\odot
\exp\big(s_{\phi}(\mathbf{x}_1)\big)
+
t_{\phi}(\mathbf{x}_1),
\end{align}
where $s_{\phi}(\cdot)$ and $t_{\phi}(\cdot)$ are neural networks that output scale and translation parameters. This structure produces a triangular Jacobian, allowing the log-determinant to be computed efficiently:
\begin{equation}
\log
\left|
\det
\frac{\partial \mathbf{y}}{\partial \mathbf{x}}
\right|
=
\sum_j s_{\phi}(\mathbf{x}_1)_j .
\end{equation}
Because of this triangular Jacobian structure, RealNVP scales efficiently with both the input dimensionality and the number of coupling layers.

For anomaly detection, the flow assigns lower likelihood to samples that deviate from the learned nominal distribution. The corresponding negative log-likelihood (NLL) anomaly score is
\begin{equation}
\mathcal{A}(\mathbf{x})
=
-\log p_X(\mathbf{x}),
\end{equation}
where higher values indicate lower probability under the learned density and therefore greater deviation from nominal behavior. In our experiments, the RealNVP model is trained and evaluated with respect to its ability to detect anomalies, using average precision, computed as the area under the precision-recall curve (PR-AUC), and area under the receiver operator characteristics (ROC-AUC) curve.

\subsection{Hybrid Latent-Structural Fusion (HLSF)}

Tensor factorization and normalizing flows capture complementary notions of normality. CP-APR models higher-order relational structure through a low-rank Poisson factorization, whereas normalizing flows perform flexible density estimation over continuous representations. To the best of our knowledge, prior work has not combined these approaches for cyber anomaly detection. We therefore propose Hybrid Latent-Structural Fusion (HLSF), which applies flow-based density estimation to latent representations derived from CP-APR and combines the resulting anomaly scores.

HLSF first uses the fitted CP-APR model to construct an event-level latent representation. Given the fitted model
\begin{equation}
\Theta
=
\{A^{(1)},\ldots,A^{(D)},\gamma\},
\end{equation}
each observed tensor entry indexed by
\begin{equation}
    \mathbf{i} = (i_1,\ldots,i_D)
\end{equation}
is mapped to a feature vector by concatenating the corresponding rows of the learned factor matrices:
\begin{equation}
\tilde{\mathbf{x}}(\mathbf{i})
=
\Big[
\log(A^{(1)}_{i_1,:}+\epsilon),
\ldots,
\log(A^{(D)}_{i_D,:}+\epsilon)
\Big]
\in
\mathbb{R}^{DR}.
\label{eq:latent_event_representation}
\end{equation}
Here, $\epsilon > 0$ is a small constant used to avoid undefined logarithms. The resulting feature vector has dimensionality $D \times R$, because $R$ latent factor values are extracted from each of the $D$ tensor modes.

Returning to the example in Figure~\ref{bin_tensor}, a $D=3$ entry representing user $i_1$ authenticating from source device $i_2$ to destination device $i_3$ is represented by concatenating the CP-APR latent factor row for that user, the latent factor row for that source device, and the latent factor row for that destination device. Thus, the raw sparse tensor entry $\mathcal{X}_{i_1,i_2,i_3}$ is transformed into a dense latent representation $\tilde{\mathbf{x}}(i_1,i_2,i_3)$ that summarizes the learned behavioral factors associated with each entity in the event.

The transformed representations are used as inputs to the RealNVP normalizing flow. The flow estimates the density
\begin{equation}
p_{\mathrm{flow}}
\left(
\tilde{\mathbf{x}}
\right)
=
p_Z
\left(
f_{\phi}
\left(
\tilde{\mathbf{x}}
\right)
\right)
\left|
\det
\left(
\frac{\partial f_{\phi}}
{\partial \tilde{\mathbf{x}}}
\right)
\right|.
\end{equation}
Unlike the Poisson likelihood, which evaluates whether an observed count is compatible with the learned low-rank tensor structure, the flow density evaluates whether the event's latent representation lies in a high- or low-density region of the learned behavioral representation space. Since the flow is trained on compact latent representations rather than directly on the full tensor space, the additional computational cost is substantially smaller than performing density estimation over all possible tensor entries.

The flow-based anomaly score for an event is the negative log-likelihood of its latent representation:
\begin{equation}
s_{\mathrm{flow}}(\mathbf{i})
=
-\log
p_{\mathrm{flow}}
\big(
\tilde{\mathbf{x}}(\mathbf{i})
\big).
\label{eq:flow_score}
\end{equation}
The CP-APR score in Equation~\eqref{eq:cp_score} and the flow score in Equation~\eqref{eq:flow_score} are first z-score normalized using validation data. The final HLSF anomaly score is then computed as
\begin{equation}
s_{\text{hybrid}}(\mathbf{i})
=
\alpha
s_{\text{CP}}(\mathbf{i})
+
(1-\alpha)
s_{\text{flow}}(\mathbf{i}),
\qquad
\alpha \in [0,1].
\label{eq:hybrid_score}
\end{equation}
The fusion weight $\alpha$ controls the relative contribution of the structural CP-APR score and the latent density score. We selected $\alpha$ using validation data by evaluating several candidate values. Because $\alpha=0.5$ consistently achieved strong performance across datasets, equal weighting was used in all reported experiments.

The two components therefore capture distinct but complementary signals. CP-APR identifies events that deviate from the learned relational structure of the sparse tensor, while the normalizing flow identifies events whose latent representations occupy low-density regions of the learned behavioral manifold. By combining these sources of information, HLSF can detect both structurally anomalous events and events that appear structurally plausible but exhibit unusual latent behavioral characteristics. In the following section, we compare CP-APR, RealNVP, and HLSF across the experimental datasets.

\section{Dataset and Experimental Setup}

In this section, we describe the datasets used to evaluate the proposed method and the experimental setup used to compare CP-APR, RealNVP, and HLSF.

\subsection{Dataset \& Tensors}

\begin{table}[t!]
\caption{LANL Authentication dataset attribute counts and selected days for dataset splits. The table summarizes the number of users, source computers, destination computers, benign events, anomalous events, and day ranges used for training, validation, combined training-validation, and testing.}
\label{table:set_table}
\resizebox{0.85\columnwidth}{!}{
\centering
\begin{tabular}{@{}lcccccc@{}}
\toprule
\textbf{Set} & \textbf{User} & \textbf{Source} & \textbf{Destination} & \textbf{Benign Events} & \textbf{Anomalous Events} & \textbf{Days} \\
\midrule
Train & 11,118 & 14,705 & 4,698 & 166,712,680 & 0 & 1-48 \\
Validation & 9,181 & 10,778 & 3,508 & 28,013,171 & 0 & 49-56 \\
Train + Validation & 11,260 & 15,055 & 4,796 & 194,841,640 & 0 & 1-56 \\
Test & 10,165 & 12,526 & 4,176 & 91,547,561 & 179 & 57-82 \\
\bottomrule
\end{tabular}
}
\end{table}

\begin{table}[t!]
\caption{Tensor statistics for each LANL dataset, including tensor dimensions, sparsity, decomposed rank, and benign and anomalous event counts as tensor entries $\mathcal{X}_{\textbf{i}}$ in the test set. The LANL tensors span large, high-dimensional spaces, but only a very small fraction of entries are nonzero, making sparse tensor representations memory efficient despite the large nominal tensor sizes. The event counts also illustrate the severe class imbalance in the test set, where anomalous events are rare relative to benign events, creating a needle-in-a-haystack detection setting.}
\label{table_tensor_statistics}
\resizebox{\columnwidth}{!}{
\centering
\begin{tabular}{@{}lccccc@{}}
\toprule
\textbf{LANL Dataset} & \textbf{Dimensions} & \textbf{\% Non-Zero} & \textbf{Decomposed Rank} & \textbf{Benign Count} & \textbf{Anomaly Count} \\
\midrule
US & $11,260 \times 15,055$ & $2.57 \times 10^{-4}$ & 20 & 31,241 & 76 \\
UD & $11,260 \times 4,796$ & $1.51 \times 10^{-3}$ & 20 & 69,596 & 117 \\
USD & $11,260 \times 15,055 \times 4,796 \times 2$ & $1.02 \times 10^{-7}$ & 4 & 125,166 & 119 \\
USDs & $11,260 \times 15,055 \times 4,796 \times 2$ & $1.02 \times 10^{-7}$ & 4 & 125,166 & 119 \\
USDHs & $11,260 \times 15,055 \times 4,796 \times 24 \times 2$ & $3.04 \times 10^{-8}$ & 5 & 955,808 & 137 \\
USDHDs & $11,260 \times 15,055 \times 4,796 \times 24 \times 7 \times 2$ & $1.60 \times 10^{-8}$ & 45 & 3,513,527 & 138 \\
\bottomrule
\end{tabular}
}
\end{table}

We evaluate the methods on six cyber tensors constructed from the Unified Host and Network Dataset introduced by \cite{Turcotte2017UnifiedHA}. The dataset contains authentication activity collected at LANL and includes red-team activity in which user credentials were compromised. In this setting, an anomalous event corresponds to a successful authentication by a red-team actor using a compromised user account. This produces a highly imbalanced anomaly-detection problem in which anomalous events are rare relative to benign authentication events.

The authentication records are represented as sparse, nonnegative count tensors. A transaction dataset can be tensorized as
\[
\mathcal{X} \in \mathbb{R}_{\ge 0}^{I_1 \times I_2 \times \cdots \times I_N},
\]
where each mode corresponds to an attribute of the event, such as \emph{user}, \emph{source device}, \emph{destination device}, \emph{hour}, \emph{day}, or authentication \emph{status}. The goal of anomaly detection is to assign an anomaly score to tensor entries, where larger scores indicate more suspicious behavior. In this paper, tensor entries are binary: an entry value of 1 indicates that at least one corresponding event occurred, while an entry value of 0 indicates that the event combination was not observed.

The six tensors used in the experiments increase in dimensionality and behavioral specificity. The \textit{User--Source} (US) tensor records whether a user authenticated from a source device. The \textit{User--Destination} (UD) tensor records whether a user authenticated to a destination device. The \textit{User--Source--Destination} (USD) tensor captures the joint relationship among users, source devices, and destination devices. The \textit{User--Source--Destination--status} (USDs) tensor additionally includes whether the authentication attempt succeeded or failed. The \textit{User--Source--Destination--Hour--status} (USDHs) tensor further incorporates the hour of day, and the \textit{User--Source--Destination--Hour--Day--status} (USDHDs) tensor includes both hour of day and day of week. Here, \textit{User} denotes the user initiating the logon, \textit{Source} denotes the device from which the authentication originated, \textit{Destination} denotes the device being authenticated to, \textit{Hour} denotes the hour of day from 0 to 23, \textit{Day} denotes the day of week, and \textit{status} denotes whether the logon attempt failed or succeeded.

Table~\ref{table:set_table} summarizes the train, validation, train-validation, and test splits used in the experiments. The training and validation periods contain only benign events, while the test period contains both benign and anomalous events. The combined train-validation period is used after model selection to fit the final CP-APR model before evaluating on the test set. These split statistics follow the setup used by Eren et al.~\citep{eren2023general}.

Table~\ref{table_tensor_statistics} reports the tensor dimensions, percent of nonzero entries, selected CP-APR rank, and the number of benign and anomalous tensor entries in the test set. Although the nominal tensor sizes are very large because the mode sizes multiply across dimensions, only a small fraction of entries are nonzero. This makes sparse tensor representations memory efficient and motivates the use of sparse tensor factorization methods. The test-set counts also show the severe class imbalance of the detection problem, where anomalous entries are rare relative to benign entries. Additional details on the CP-APR preprocessing procedure and prior analysis of these tensors are provided by Eren et al.~\citep{eren2023general}.

\subsection{Experimental Setup}

\begin{table}[t!]
\caption{RealNVP normalizing-flow hyperparameters selected for each LANL tensor dataset using Hyperband tuning.}
\label{table_realnvp_hyperparameters}
\resizebox{0.75\columnwidth}{!}{
\centering
\begin{tabular}{@{}lcccc@{}}
\toprule
\textbf{LANL Dataset} & \textbf{Learning Rate} & \textbf{Hidden Features} & \textbf{Batch Size} & \textbf{Number of Layers} \\
\midrule
US & $2.96 \times 10^{-4}$ & 164 & 256 & 9 \\
UD & $6.7 \times 10^{-4}$ & 181 & 64 & 4 \\
USD & $6.0 \times 10^{-4}$ & 171 & 256 & 10 \\
USDs & $1.0 \times 10^{-4}$ & 95 & 64 & 4 \\
USDHs & $5.3 \times 10^{-4}$ & 152 & 64 & 8 \\
USDHDs & $9.2 \times 10^{-4}$ & 70 & 128 & 4 \\
\bottomrule
\end{tabular}
}
\end{table}

For CP-APR, we follow the initialization and rank-selection procedure used by Eren et al.~\citep{eren2023general}. CP-APR models are fit on the training set for ranks from 1 to 100, and the validation log-likelihood is used to select the best rank. The selected decomposed ranks for each tensor are reported in Table~\ref{table_tensor_statistics}. After selecting the rank, the training and validation sets are combined to fit the final CP-APR model. The learned latent factors are then used to estimate the Poisson rate $\lambda_i$ for each test entry, and the Poisson survival function is used to assign a p-value-based anomaly score.

For RealNVP, hyperparameters are selected using Hyperband~\citep{Li2018Hyperband}. Hyperband is computationally efficient because it adaptively allocates resources across candidate configurations and terminates poorly performing configurations early. It combines random search with successive halving, allowing many configurations to be evaluated with small training budgets before increasing resources for the most promising configurations~\citep{Li2018Hyperband}. This is well suited for normalizing flows because training can be expensive and the hyperparameter space is high-dimensional. The selected RealNVP hyperparameters for each tensor are shown in Table~\ref{table_realnvp_hyperparameters}.

During tuning, the RealNVP model is trained on unlabeled data. Because labeled anomalies are often scarce or unavailable during training, we use a semi-supervised model-selection procedure in which the labeled test period is partitioned into validation and fully held-out test subsets~\citep{CHEN2025122463,zhousemisup23}. The split is performed temporally so that the validation subset contains anomalies for model selection and the held-out test subset contains additional anomalies for final evaluation. Hyperparameter optimization maximizes average precision (AP), equivalent to the area under the precision-recall curve (PR-AUC), on the validation subset. The best configuration for each dataset is then evaluated on the fully held-out labeled test subset.

The same evaluation protocol is used to compare CP-APR, RealNVP, and HLSF. CP-APR provides a structural anomaly score based on the fitted sparse Poisson tensor model, RealNVP provides a latent density score based on negative log-likelihood, and HLSF combines these two scores after normalization. This setup enables a direct comparison between tensor-structural anomaly detection, flow-based density estimation, and the proposed hybrid method.

\begin{table}[h!]
\caption{Anomaly detection performance across CP-APR, RealNVP normalizing flows, and the proposed HLSF framework. The best ROC-AUC and PR-AUC score for each tensor dataset is shown in bold. HLSF provides the strongest overall performance, achieving the best ROC-AUC on five of the six LANL tensor datasets and the best PR-AUC on four of the six datasets. The gains are most pronounced for the higher-order tensor representations, including USD, USDs, USDHs, and USDHDs, where HLSF consistently improves over both CP-APR and RealNVP. The low absolute PR-AUC values reflect the extreme rarity of anomalous events relative to benign events, underscoring the needle-in-a-haystack nature of the LANL anomaly detection task.}
\label{table_method_comparison_auc}
\resizebox{\columnwidth}{!}{
\centering
\begin{tabular}{@{}lcccccc@{}}
\toprule
\textbf{LANL Dataset} & \multicolumn{2}{c}{\textbf{CP-APR}} & \multicolumn{2}{c}{\textbf{RealNVP}} & \multicolumn{2}{c}{\textbf{HLSF (ours)}} \\
\cmidrule(lr){2-3}\cmidrule(lr){4-5}\cmidrule(l){6-7}
& \textbf{ROC-AUC} & \textbf{PR-AUC} & \textbf{ROC-AUC} & \textbf{PR-AUC} & \textbf{ROC-AUC} & \textbf{PR-AUC} \\
\midrule
US & $0.8746 \pm 0.0103$ & $0.0135 \pm 0.0018$ & $0.8111 \pm 0.0220$ & $\mathbf{0.0424 \pm 0.0361}$ & $\mathbf{0.8939 \pm 0.0109}$ & $0.0409 \pm 0.0350$ \\
UD & $\mathbf{0.7727 \pm 0.0087}$ & $\mathbf{0.0042 \pm 0.0003}$ & $0.6787 \pm 0.0165$ & $0.0030 \pm 0.0001$ & $0.6895 \pm 0.0181$ & $0.0030 \pm 0.0001$ \\
USD & $0.9295 \pm 0.0143$ & $0.0092 \pm 0.0011$ & $0.9211 \pm 0.0072$ & $0.0357 \pm 0.0220$ & $\mathbf{0.9709 \pm 0.0012}$ & $\mathbf{0.0573 \pm 0.0273}$ \\
USDs & $0.9758 \pm 0.0085$ & $0.0045 \pm 0.0005$ & $0.9732 \pm 0.0061$ & $0.0399 \pm 0.0281$ & $\mathbf{0.9889 \pm 0.0012}$ & $\mathbf{0.0532 \pm 0.0296}$ \\
USDHs & $0.9775 \pm 0.0075$ & $0.0041 \pm 0.0002$ & $0.9793 \pm 0.0044$ & $0.0235 \pm 0.0159$ & $\mathbf{0.9901 \pm 0.0009}$ & $\mathbf{0.0458 \pm 0.0312}$ \\
USDHDs & $0.9847 \pm 0.0059$ & $0.0018 \pm 0.0001$ & $0.9811 \pm 0.0053$ & $0.0225 \pm 0.0254$ & $\mathbf{0.9964 \pm 0.0002}$ & $\mathbf{0.0502 \pm 0.0448}$ \\
\bottomrule
\end{tabular}
}
\end{table}

\section{Results}

\begin{figure}
    \centering
    \begin{annotatedFigure}
    {\includegraphics[width=5.5cm]{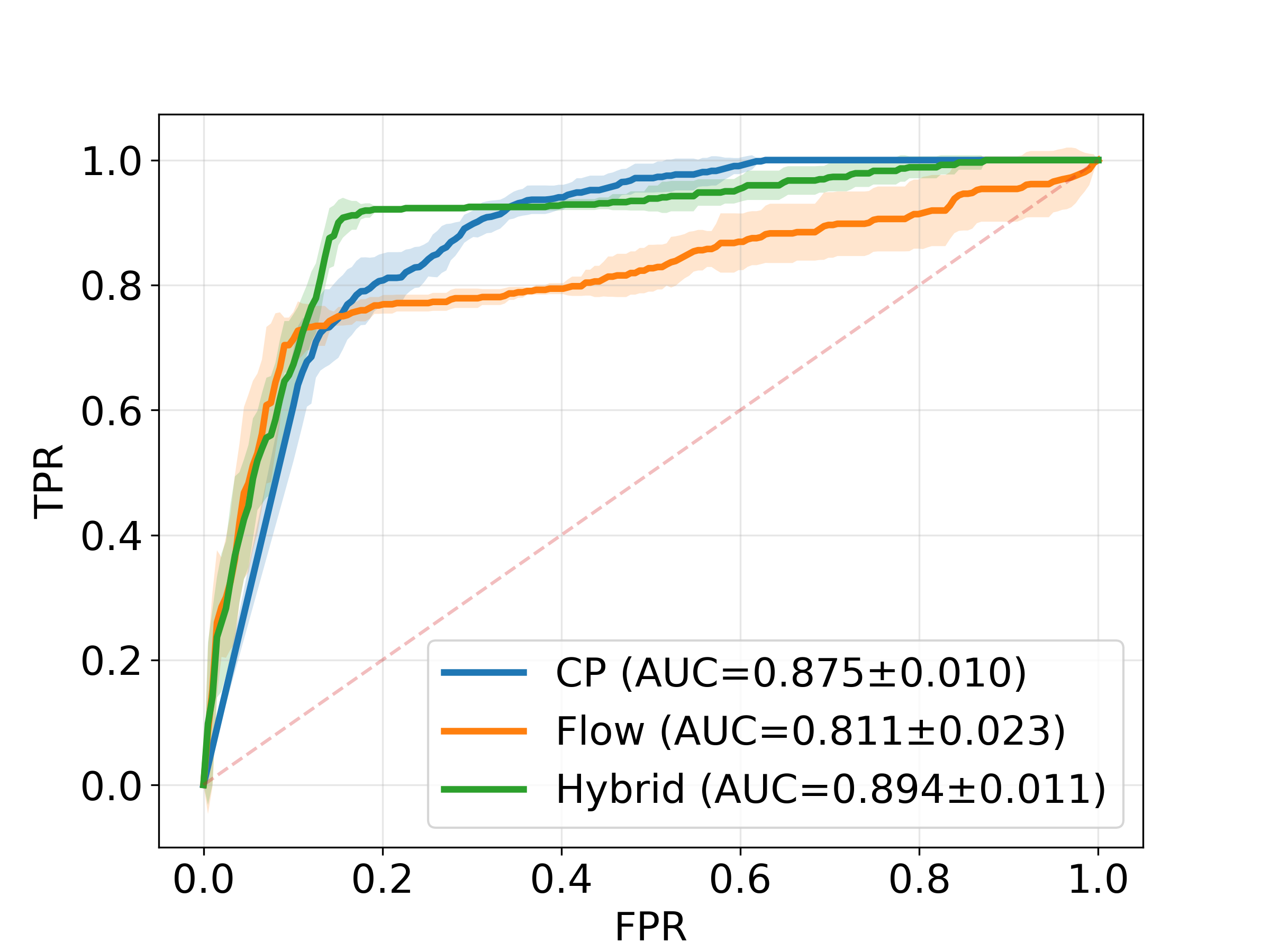}}
    \annotatedFigureText{0.4985,0.89}{black}{0.0523}{\large a} 
    \end{annotatedFigure} 
    \begin{annotatedFigure}
    {\includegraphics[width=5.5cm]{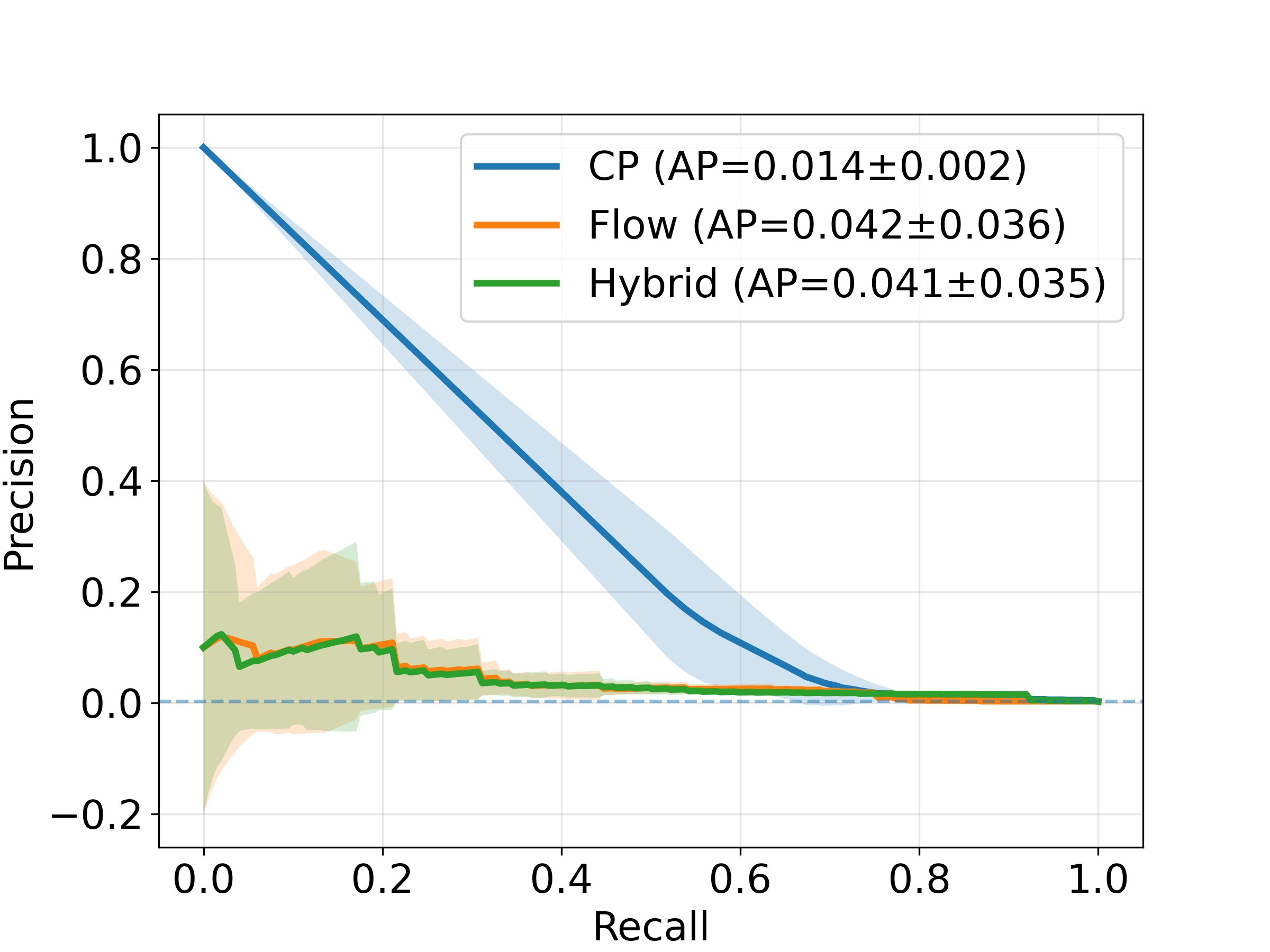}}
    \annotatedFigureText{0.4985,0.89}{black}{0.0523}{\large b} 
    \end{annotatedFigure} 
    \begin{annotatedFigure}
    {\includegraphics[width=5.5cm]{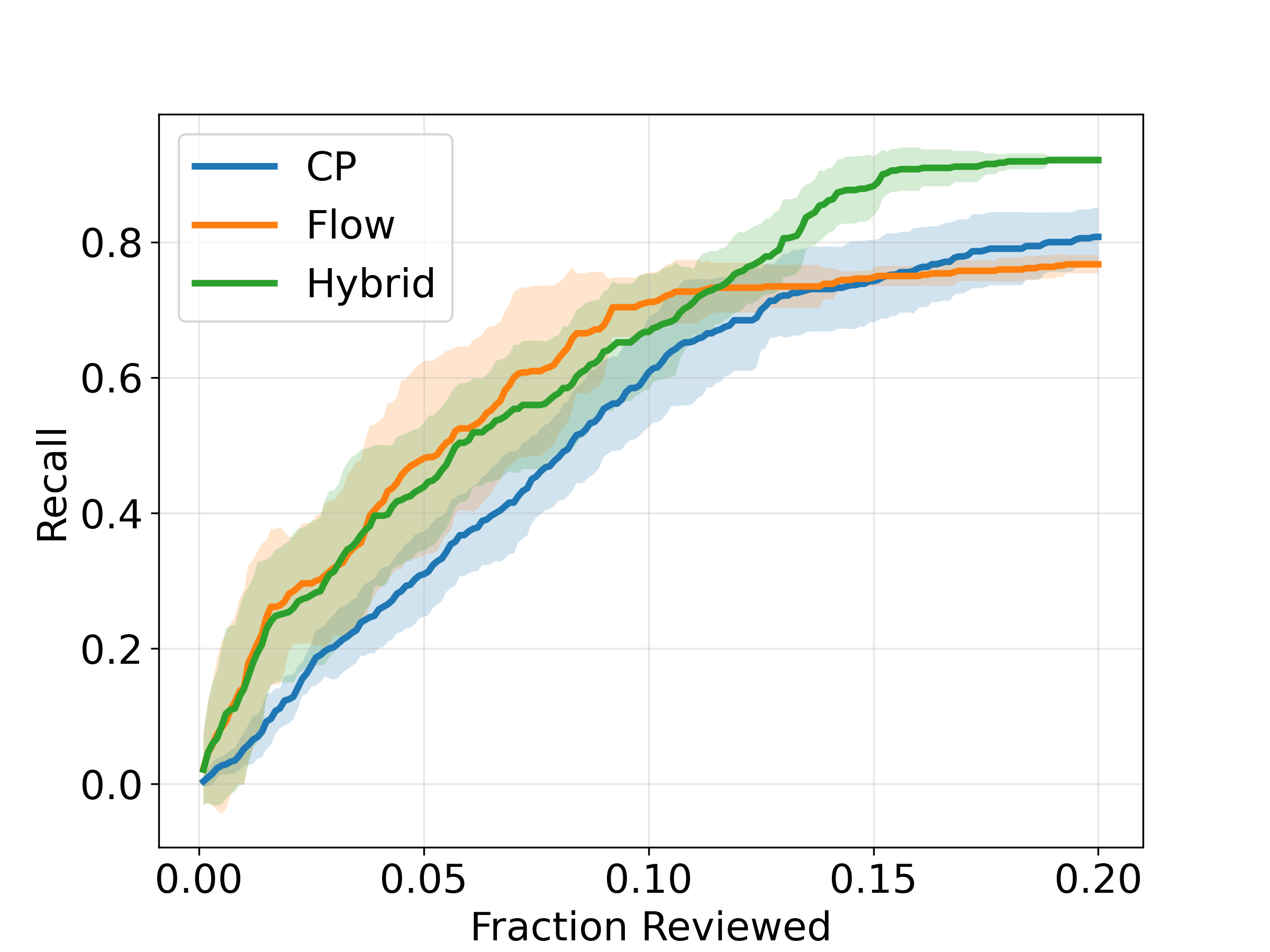}}
    \annotatedFigureText{0.4985,0.89}{black}{0.0523}{\large c} 
    \end{annotatedFigure} 
    \begin{annotatedFigure}
    {\includegraphics[width=5.5cm]{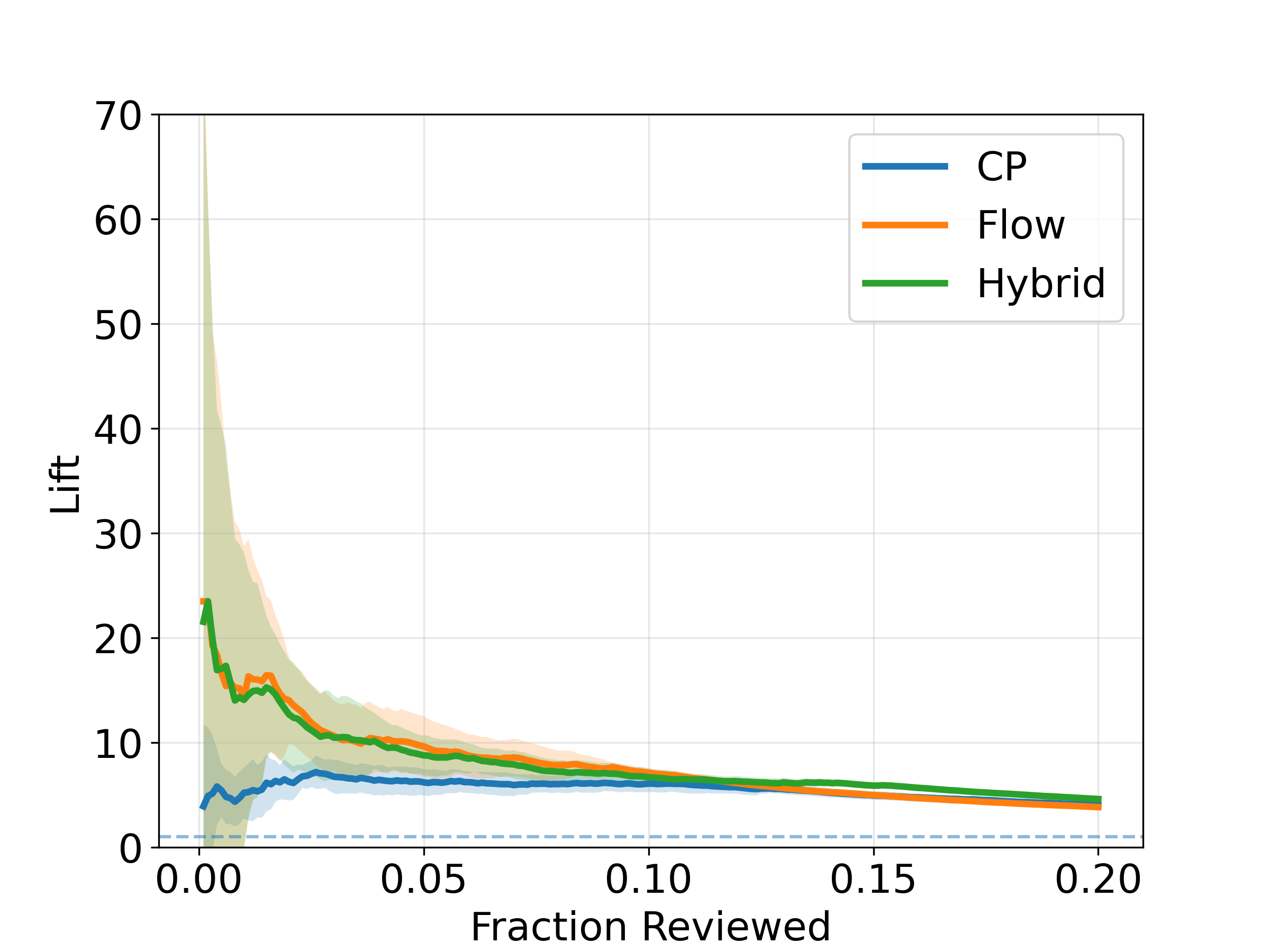}}
    \annotatedFigureText{0.4985,0.89}{black}{0.0523}{\large d} 
    \end{annotatedFigure} 
    \caption{Method comparisons using ensemble runs of the US dataset for (a) ROC, (b) precision-recall, (c) recall versus fraction of dataset reviewed, and (d) lift versus fraction reviewed.}
    \label{US}
\end{figure}

\begin{figure}
    \centering
    \begin{annotatedFigure}
    {\includegraphics[width=5.5cm]{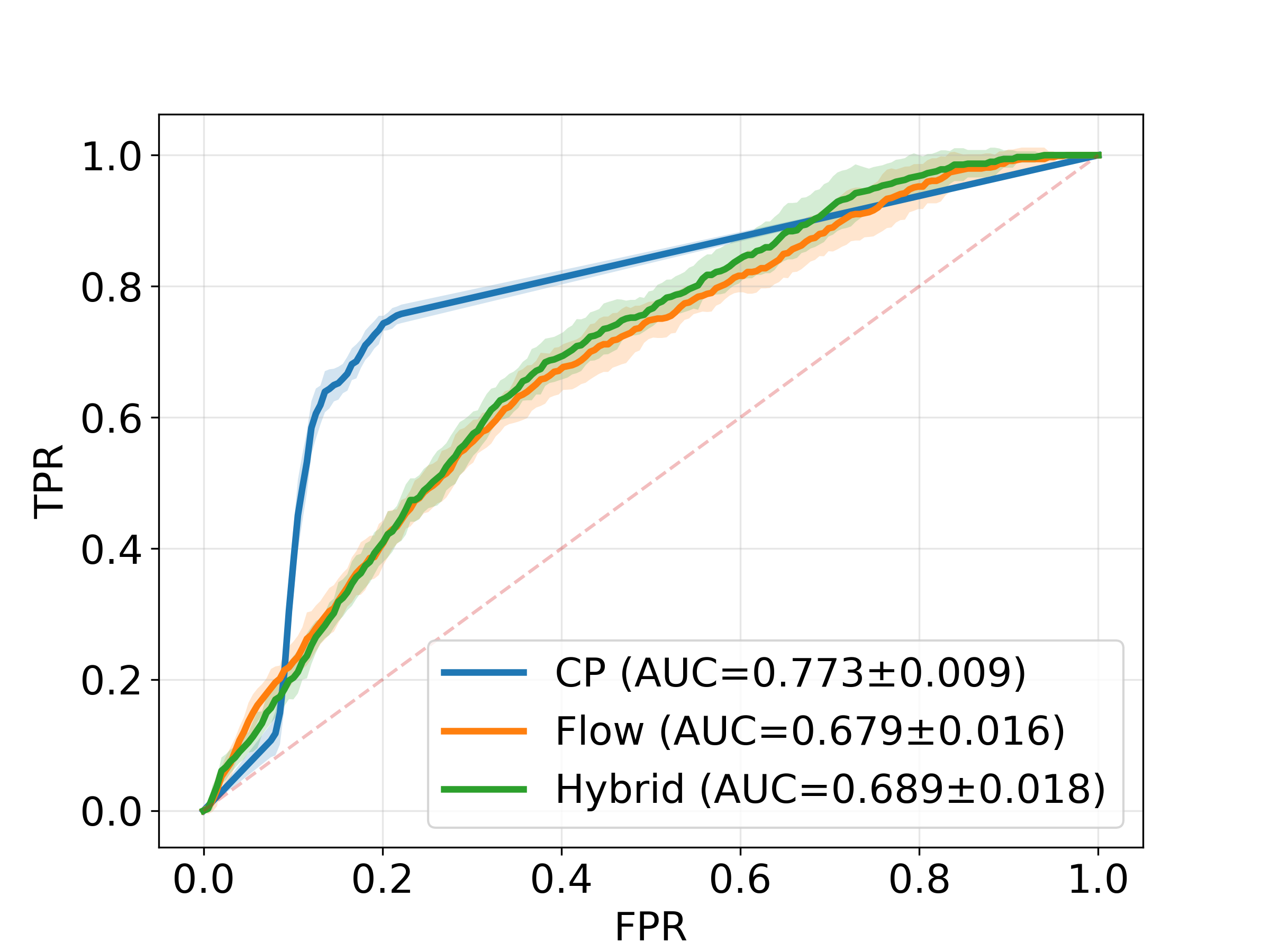}}
    \annotatedFigureText{0.4985,0.89}{black}{0.0523}{\large a} 
    \end{annotatedFigure} 
    \begin{annotatedFigure}
    {\includegraphics[width=5.5cm]{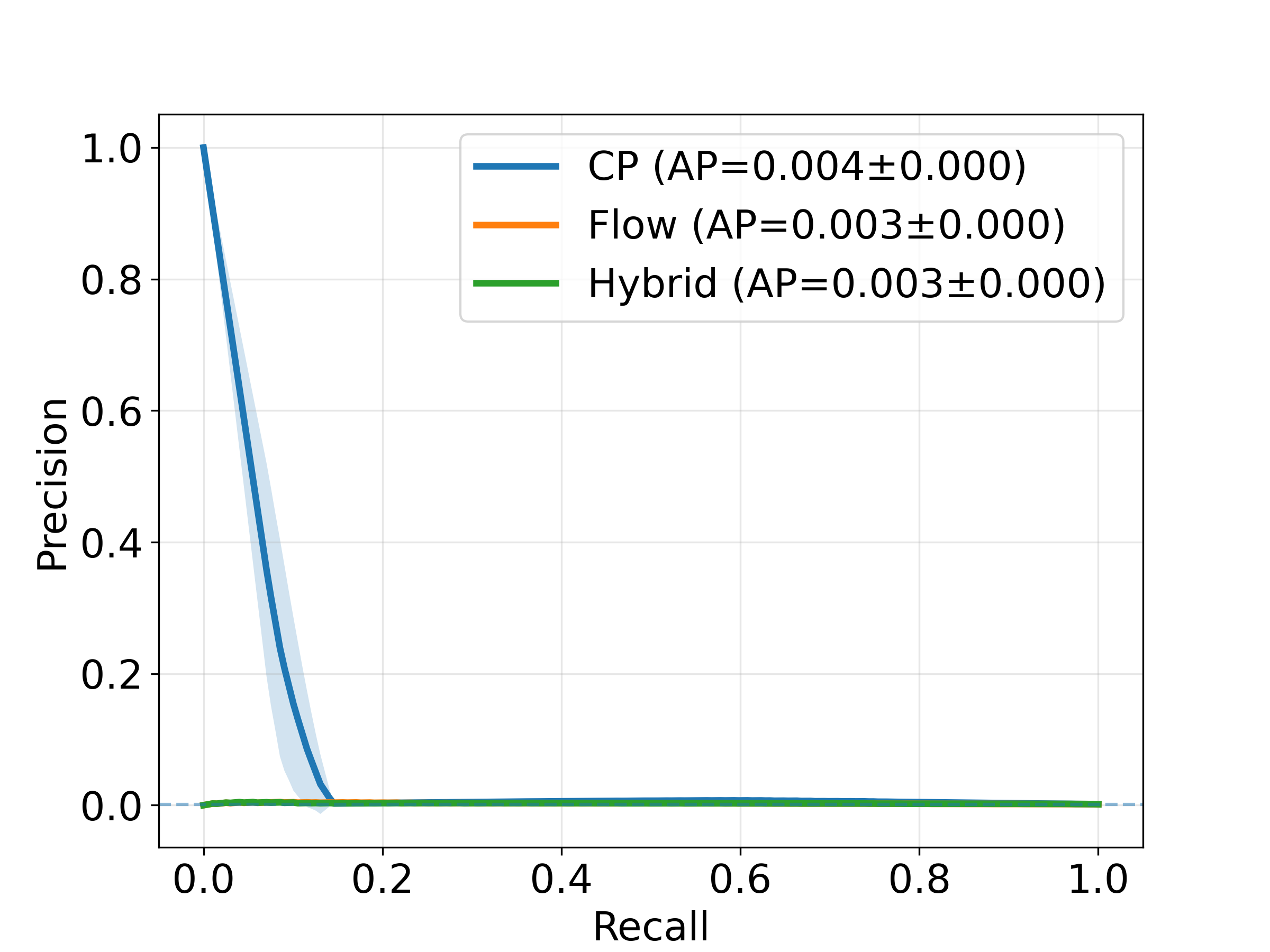}}
    \annotatedFigureText{0.4985,0.89}{black}{0.0523}{\large b} 
    \end{annotatedFigure} 
    \begin{annotatedFigure}
    {\includegraphics[width=5.5cm]{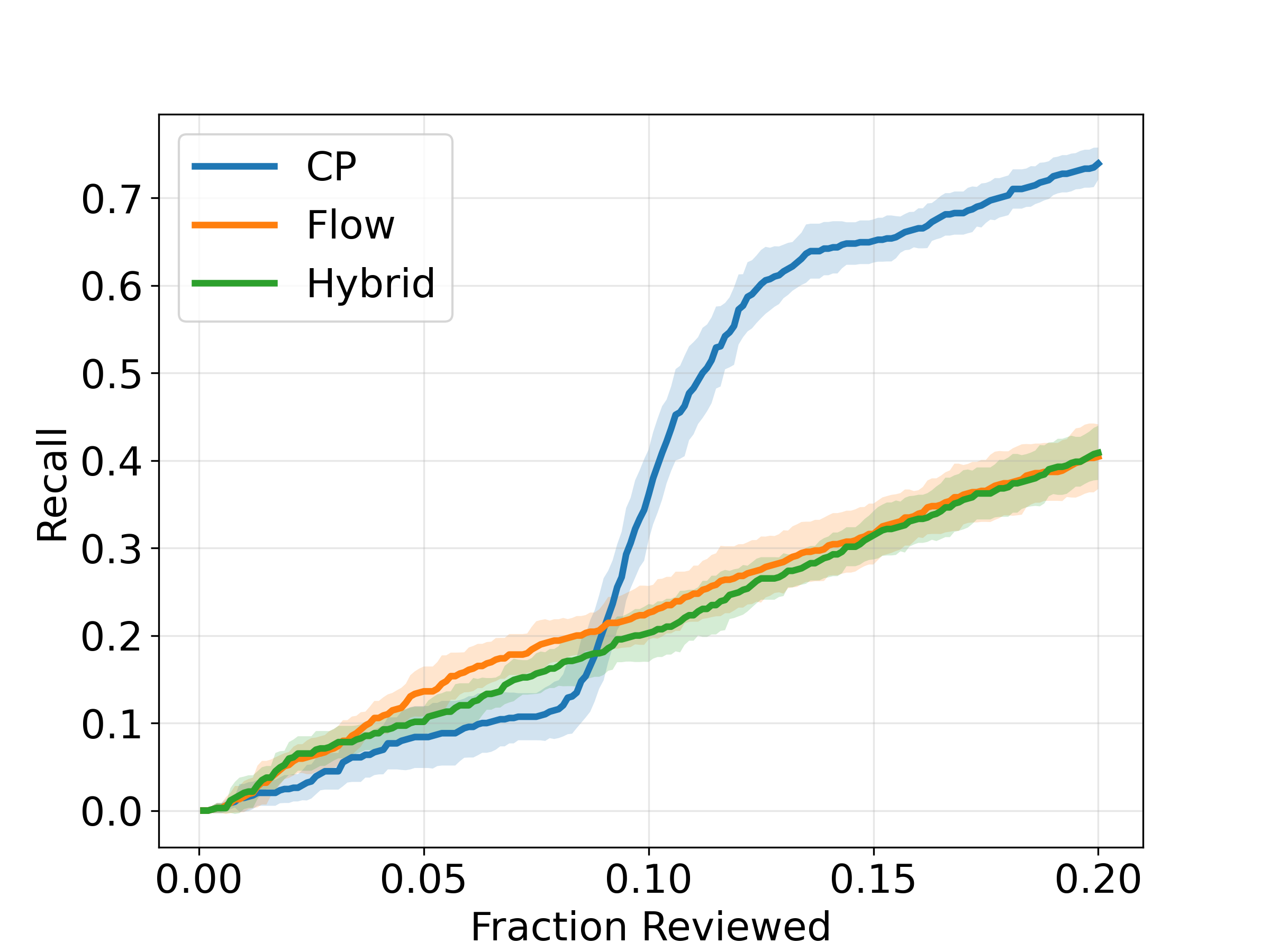}}
    \annotatedFigureText{0.4985,0.89}{black}{0.0523}{\large c} 
    \end{annotatedFigure} 
    \begin{annotatedFigure}
    {\includegraphics[width=5.5cm]{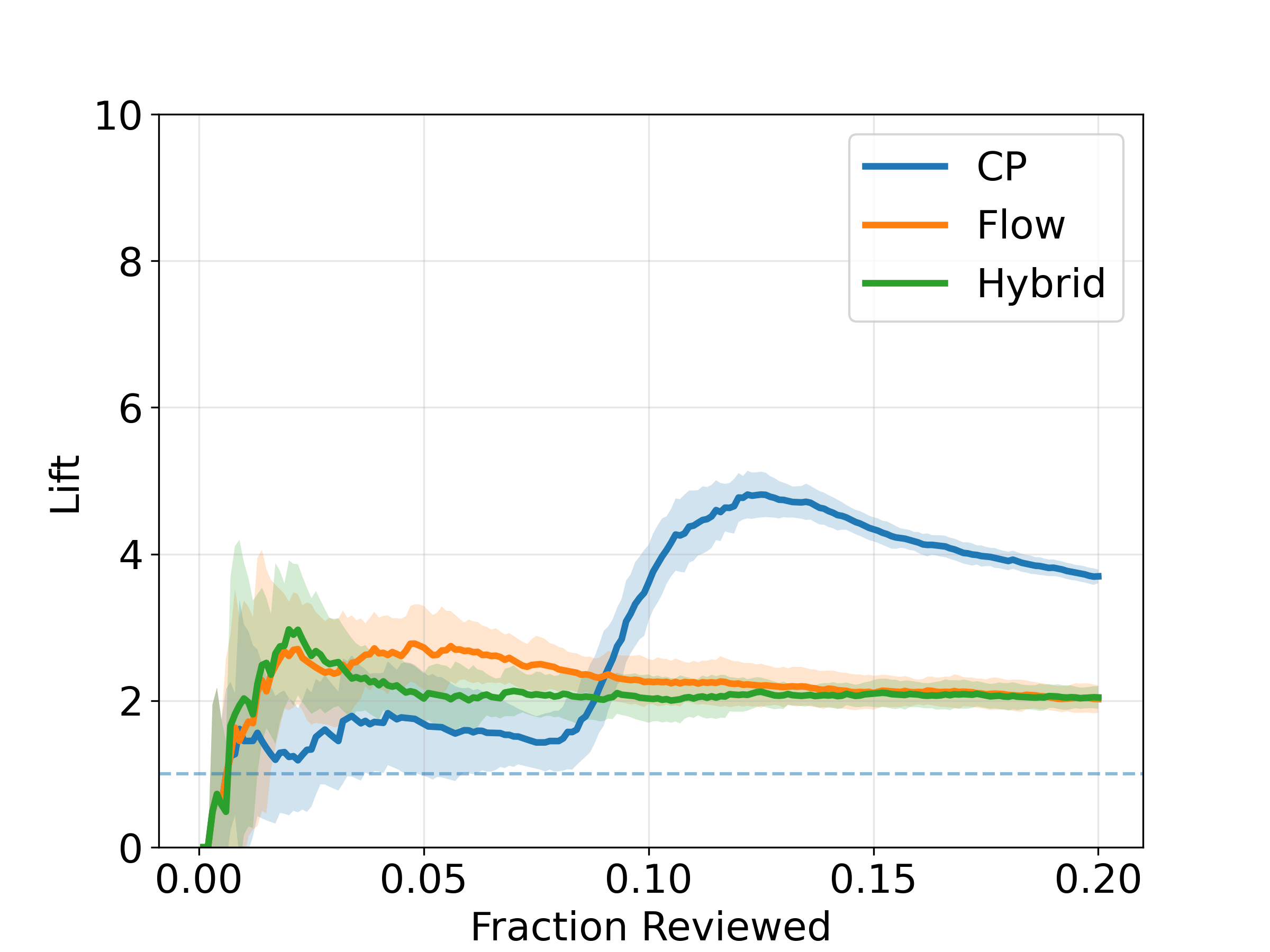}}
    \annotatedFigureText{0.4985,0.89}{black}{0.0523}{\large d} 
    \end{annotatedFigure} 
    \caption{Method comparisons using ensemble runs of the UD dataset for (a) ROC, (b) precision-recall, (c) recall versus fraction of dataset reviewed, and (d) lift versus fraction reviewed.}
    \label{UD}
\end{figure}

\begin{figure}
    \centering
    \begin{annotatedFigure}
    {\includegraphics[width=5.5cm]{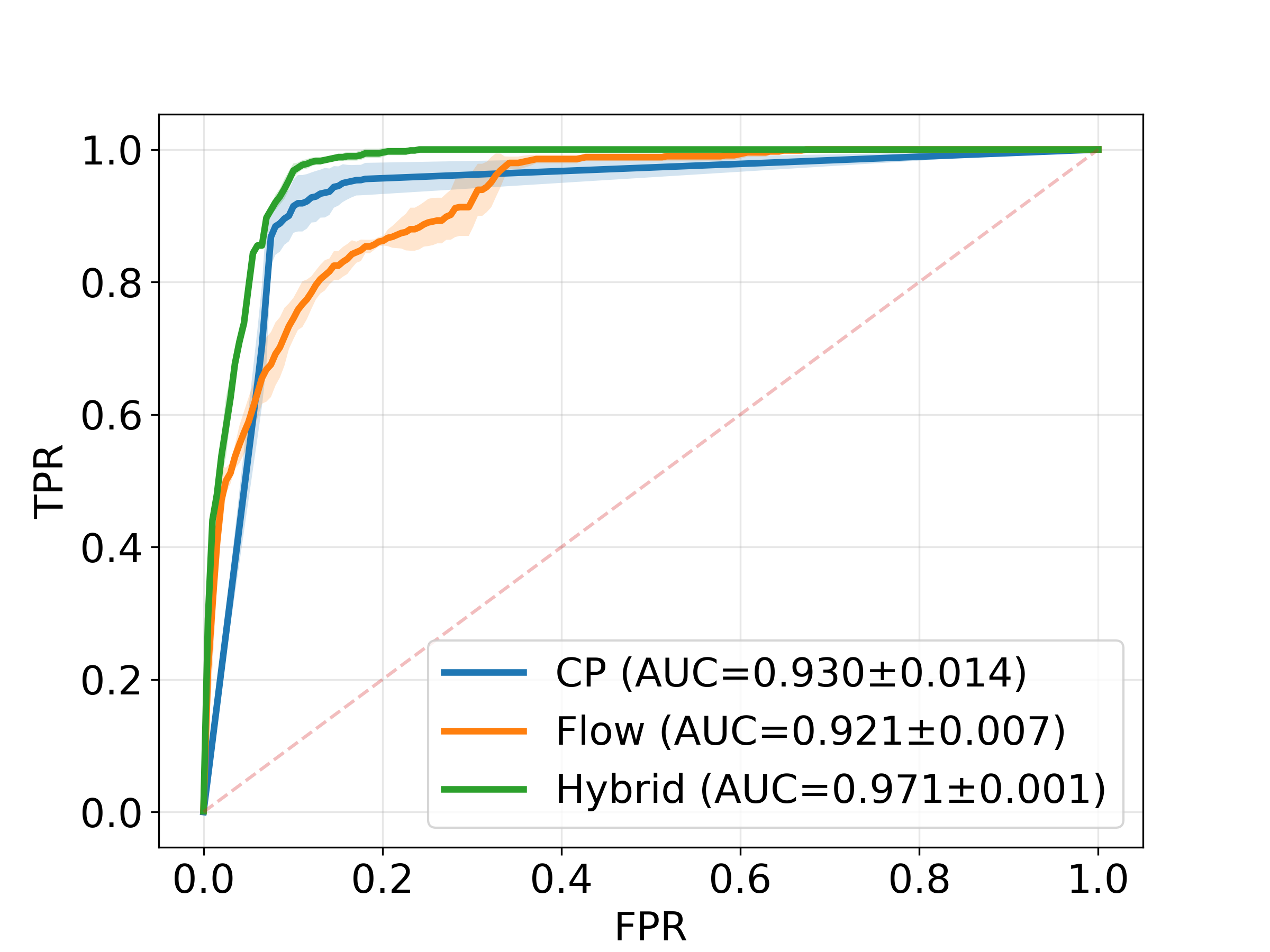}}
    \annotatedFigureText{0.4985,0.89}{black}{0.0523}{\large a} 
    \end{annotatedFigure} 
    \begin{annotatedFigure}
    {\includegraphics[width=5.5cm]{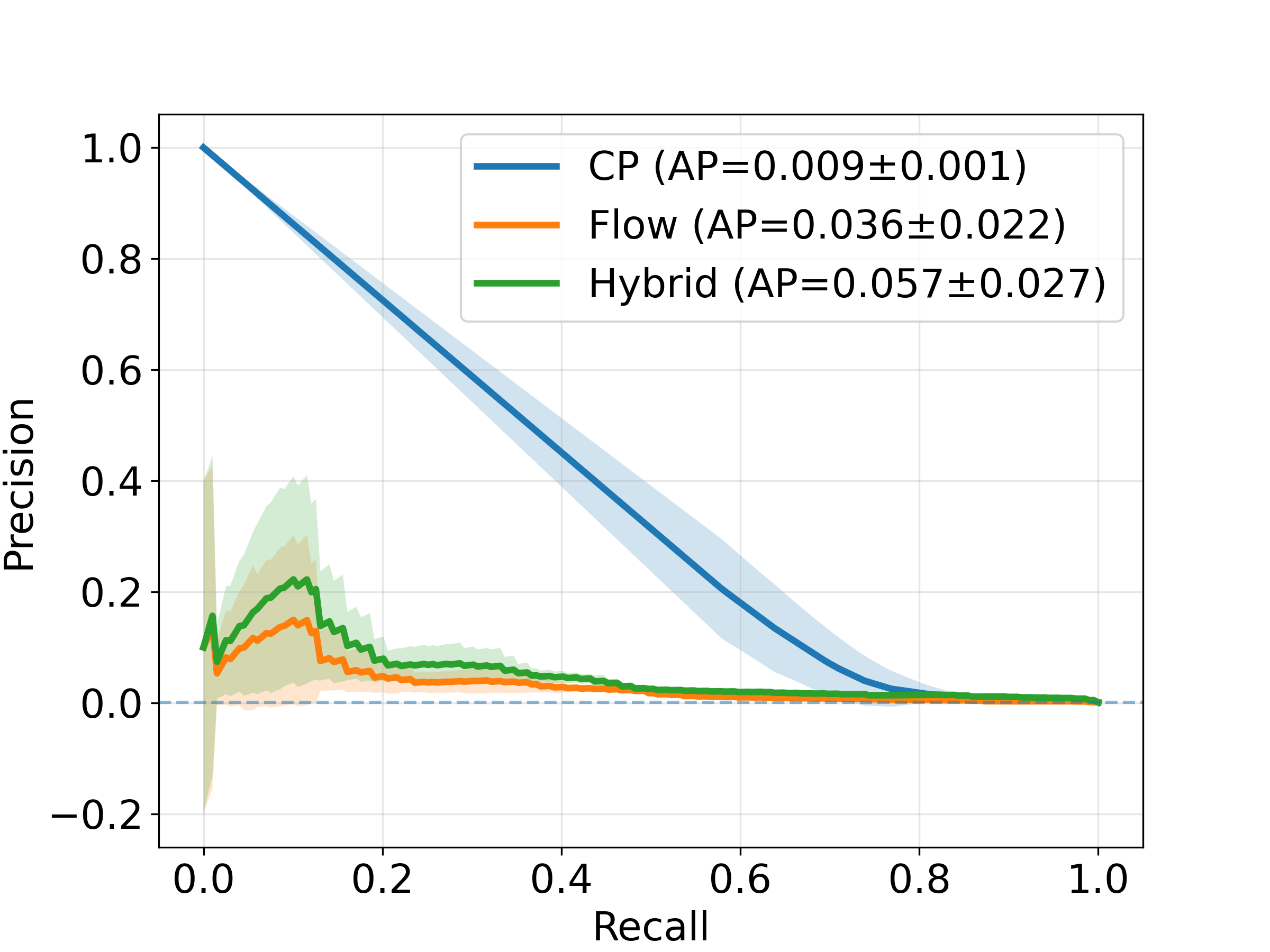}}
    \annotatedFigureText{0.4985,0.89}{black}{0.0523}{\large b} 
    \end{annotatedFigure} 
    \begin{annotatedFigure}
    {\includegraphics[width=5.5cm]{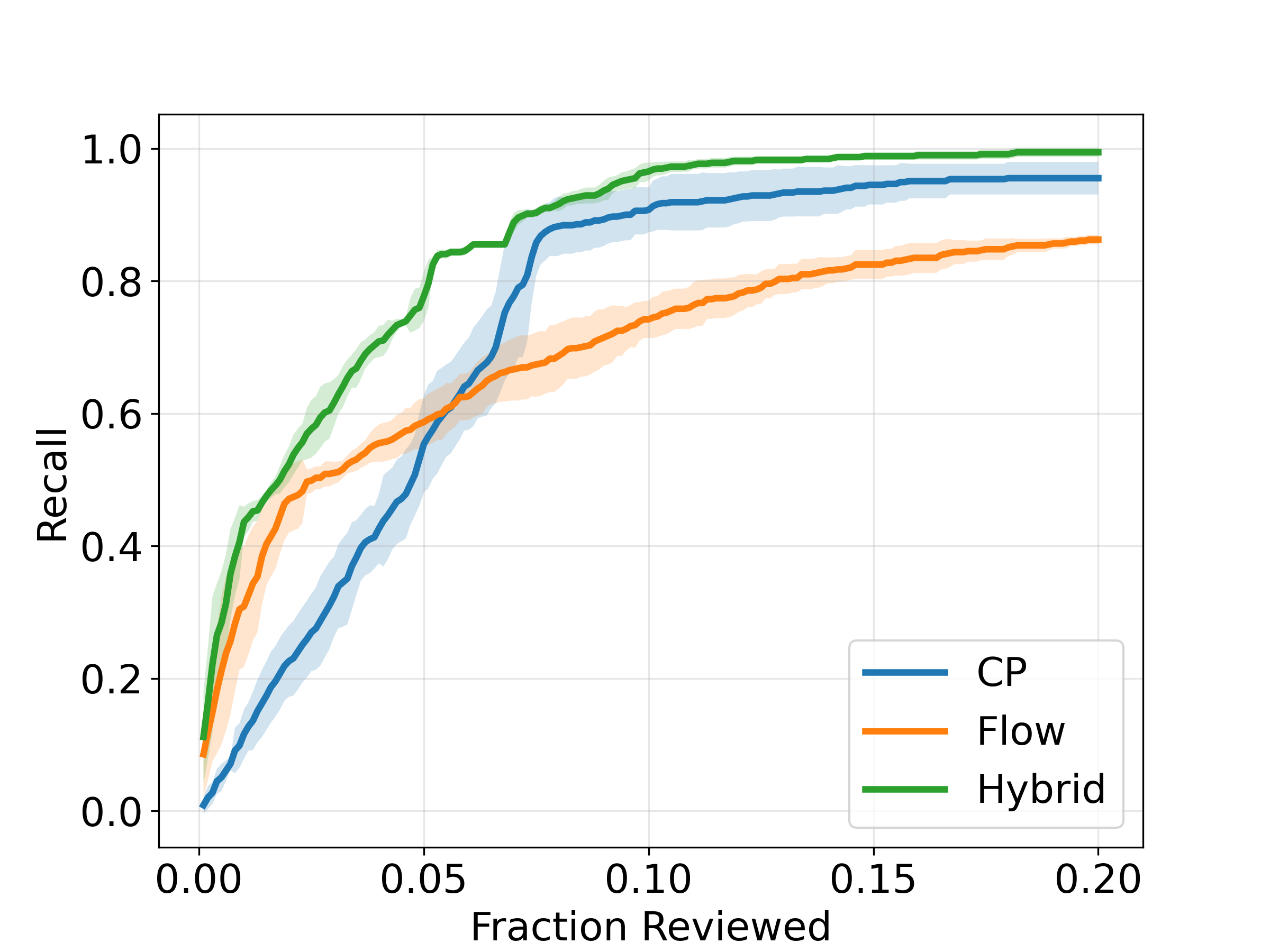}}
    \annotatedFigureText{0.4985,0.89}{black}{0.0523}{\large c} 
    \end{annotatedFigure} 
    \begin{annotatedFigure}
    {\includegraphics[width=5.5cm]{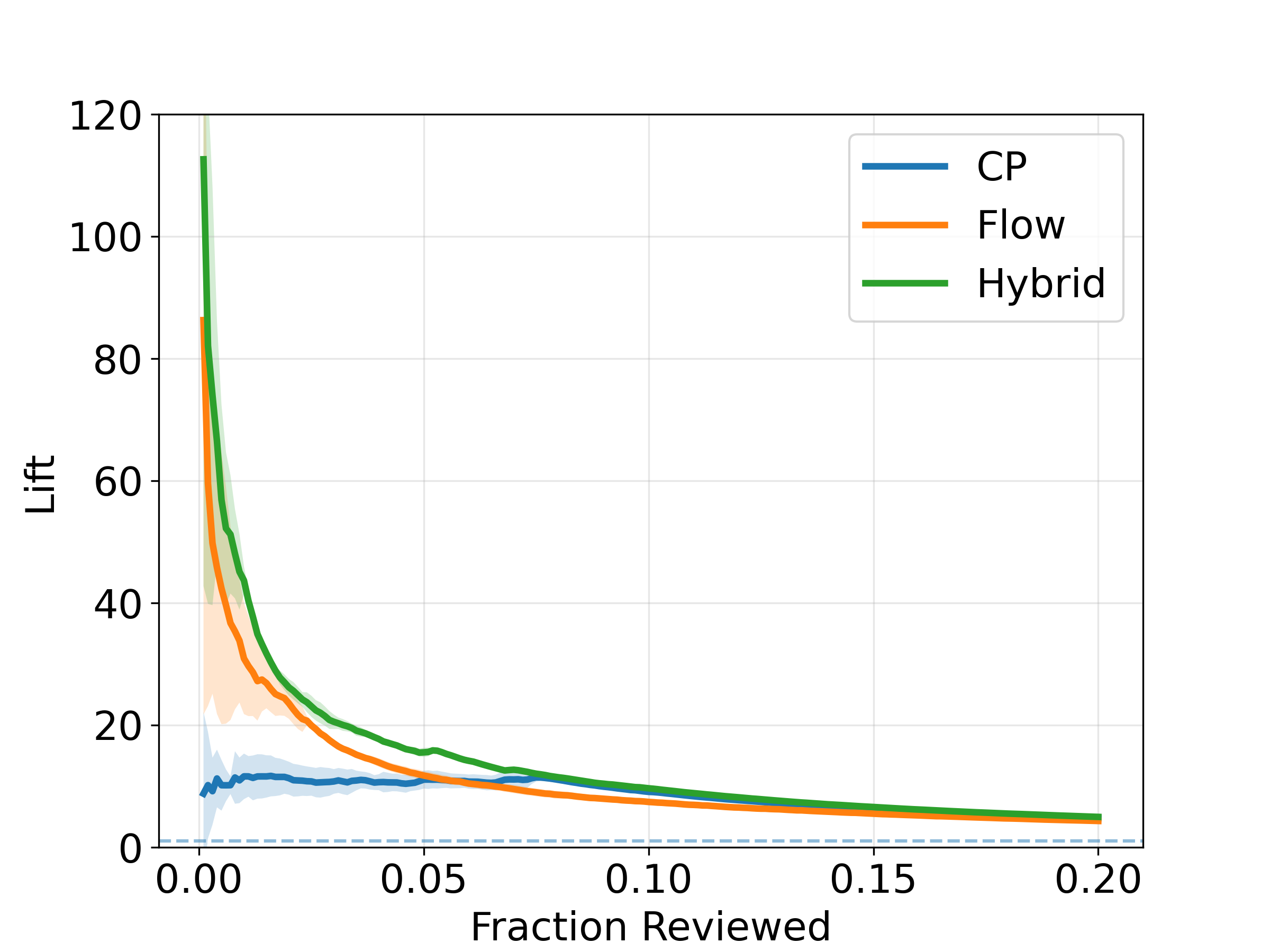}}
    \annotatedFigureText{0.4985,0.89}{black}{0.0523}{\large d} 
    \end{annotatedFigure} 
    \caption{Method comparisons using ensemble runs of the USD dataset for (a) ROC, (b) precision-recall, (c) recall versus fraction of dataset reviewed, and (d) lift versus fraction reviewed.}
    \label{USD}
\end{figure}

\begin{figure}
    \centering
    \begin{annotatedFigure}
    {\includegraphics[width=5.5cm]{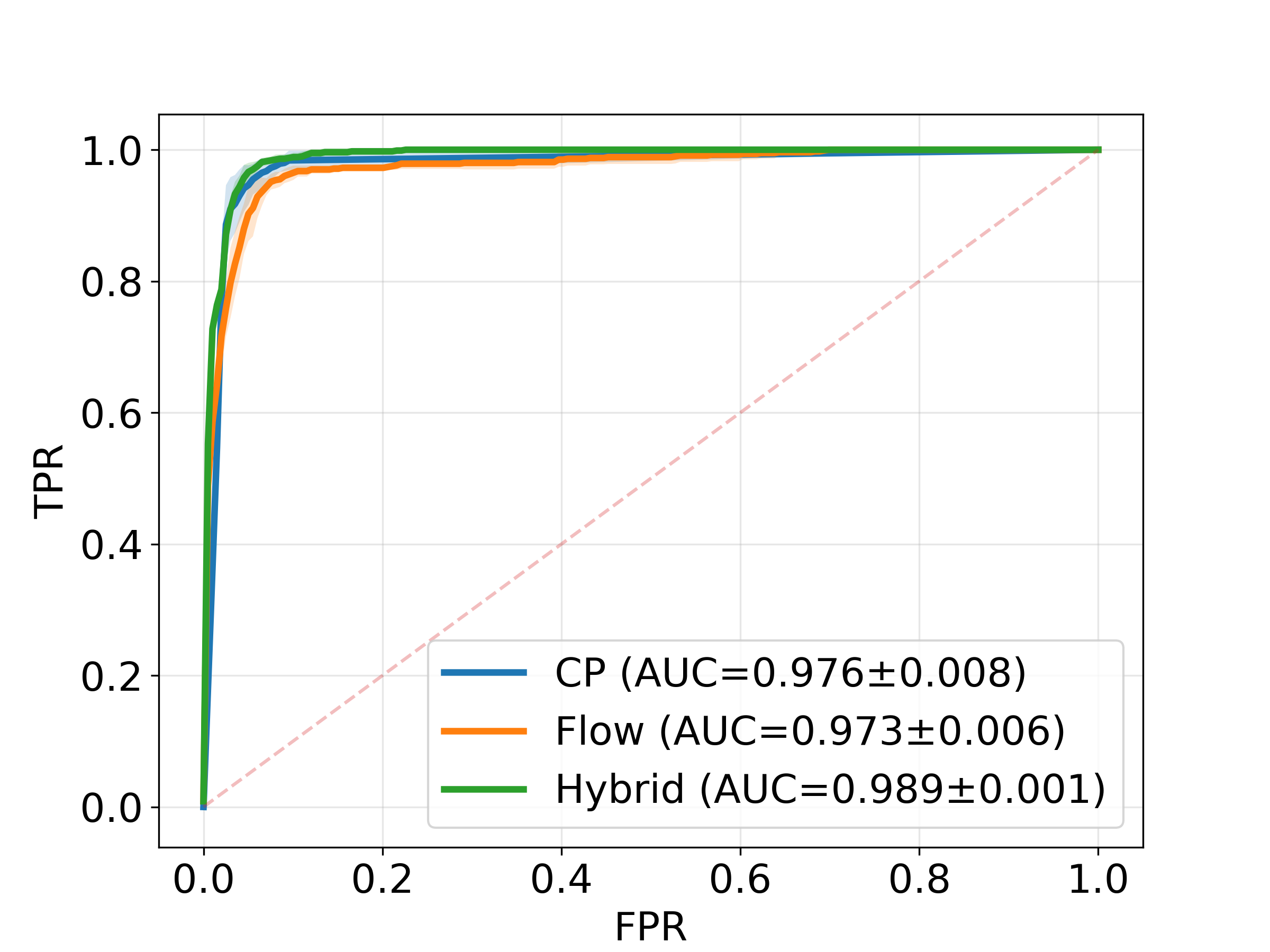}}
    \annotatedFigureText{0.4985,0.89}{black}{0.0523}{\large a} 
    \end{annotatedFigure} 
    \begin{annotatedFigure}
    {\includegraphics[width=5.5cm]{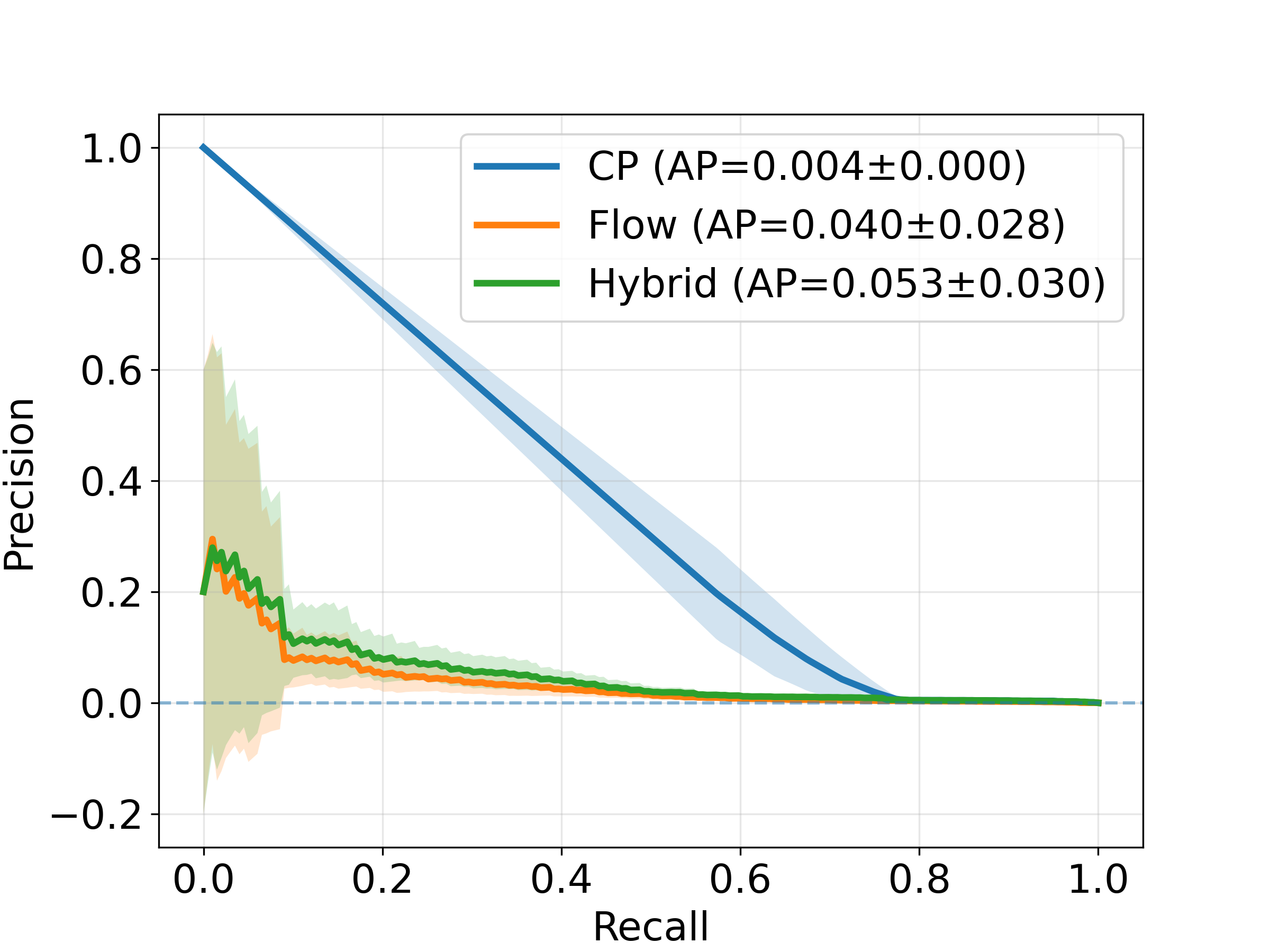}}
    \annotatedFigureText{0.4985,0.89}{black}{0.0523}{\large b} 
    \end{annotatedFigure} 
    \begin{annotatedFigure}
    {\includegraphics[width=5.5cm]{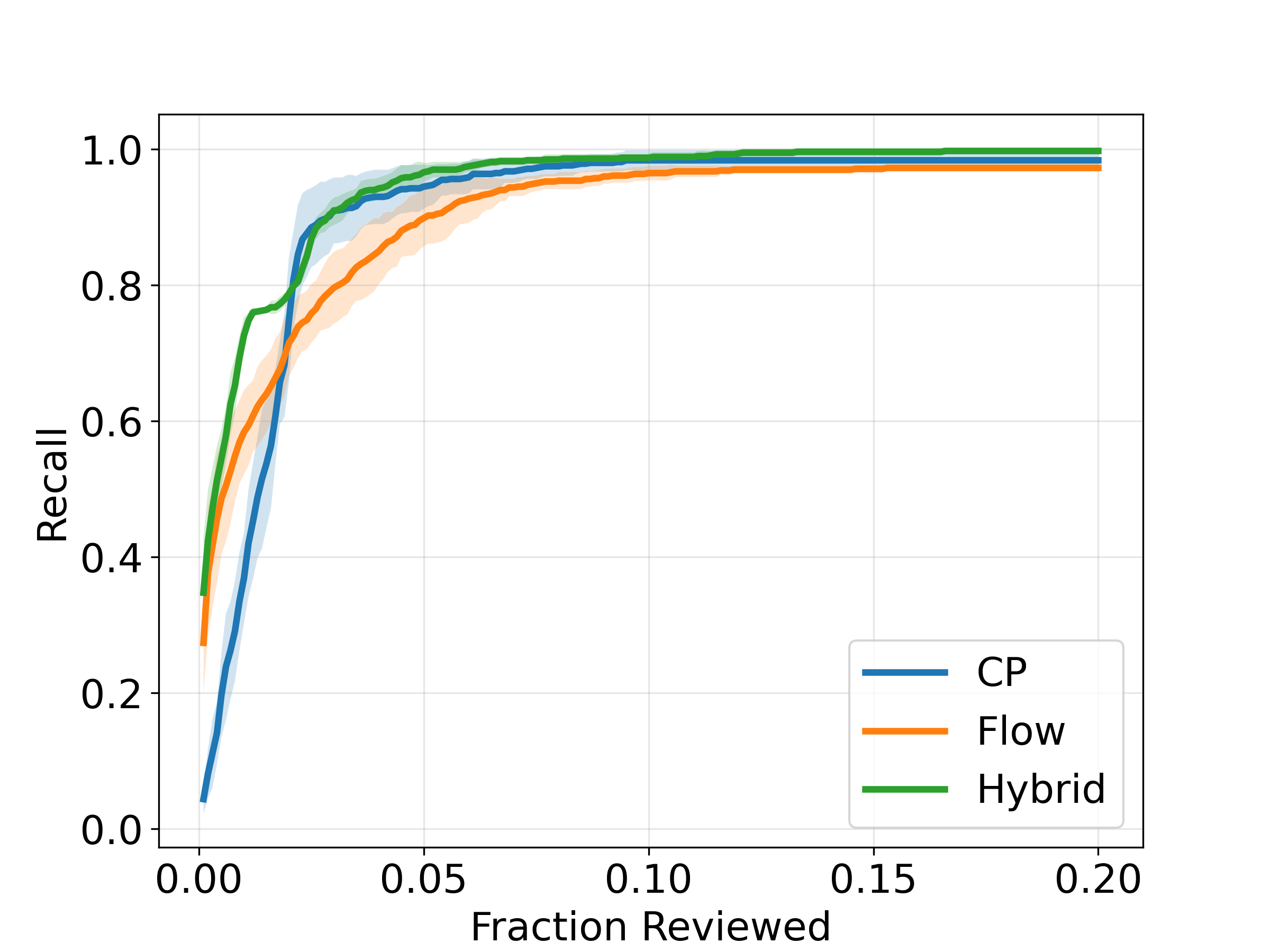}}
    \annotatedFigureText{0.4985,0.89}{black}{0.0523}{\large c} 
    \end{annotatedFigure} 
    \begin{annotatedFigure}
    {\includegraphics[width=5.5cm]{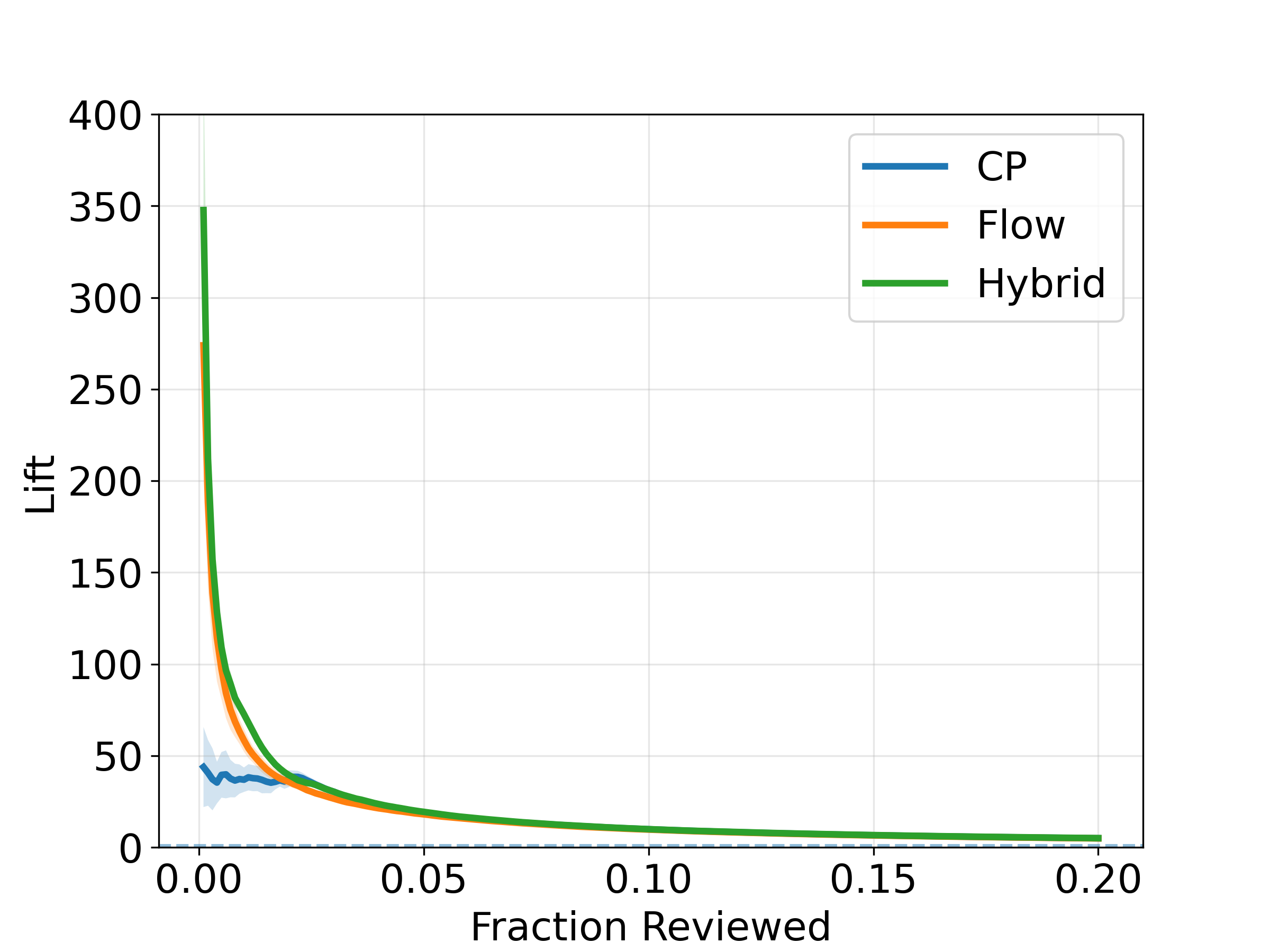}}
    \annotatedFigureText{0.4985,0.89}{black}{0.0523}{\large d} 
    \end{annotatedFigure} 
    \caption{Method comparisons using ensemble runs of the USDs dataset for (a) ROC, (b) precision-recall, (c) recall versus fraction of dataset reviewed, and (d) lift versus fraction reviewed.}
    \label{USDs}
\end{figure}

\begin{figure}
    \centering
    \begin{annotatedFigure}
    {\includegraphics[width=5.5cm]{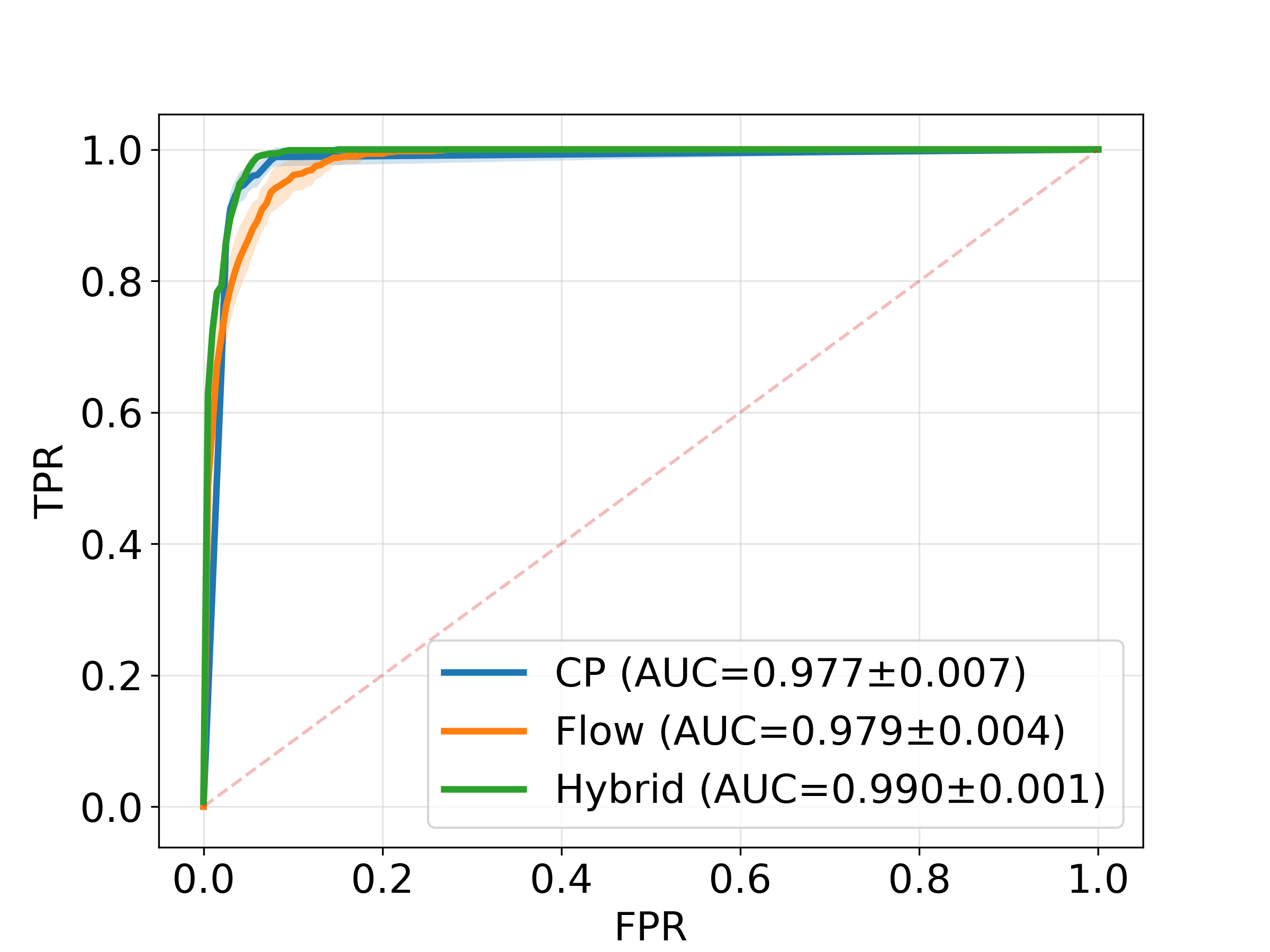}}
    \annotatedFigureText{0.4985,0.89}{black}{0.0523}{\large a} 
    \end{annotatedFigure} 
    \begin{annotatedFigure}
    {\includegraphics[width=5.5cm]{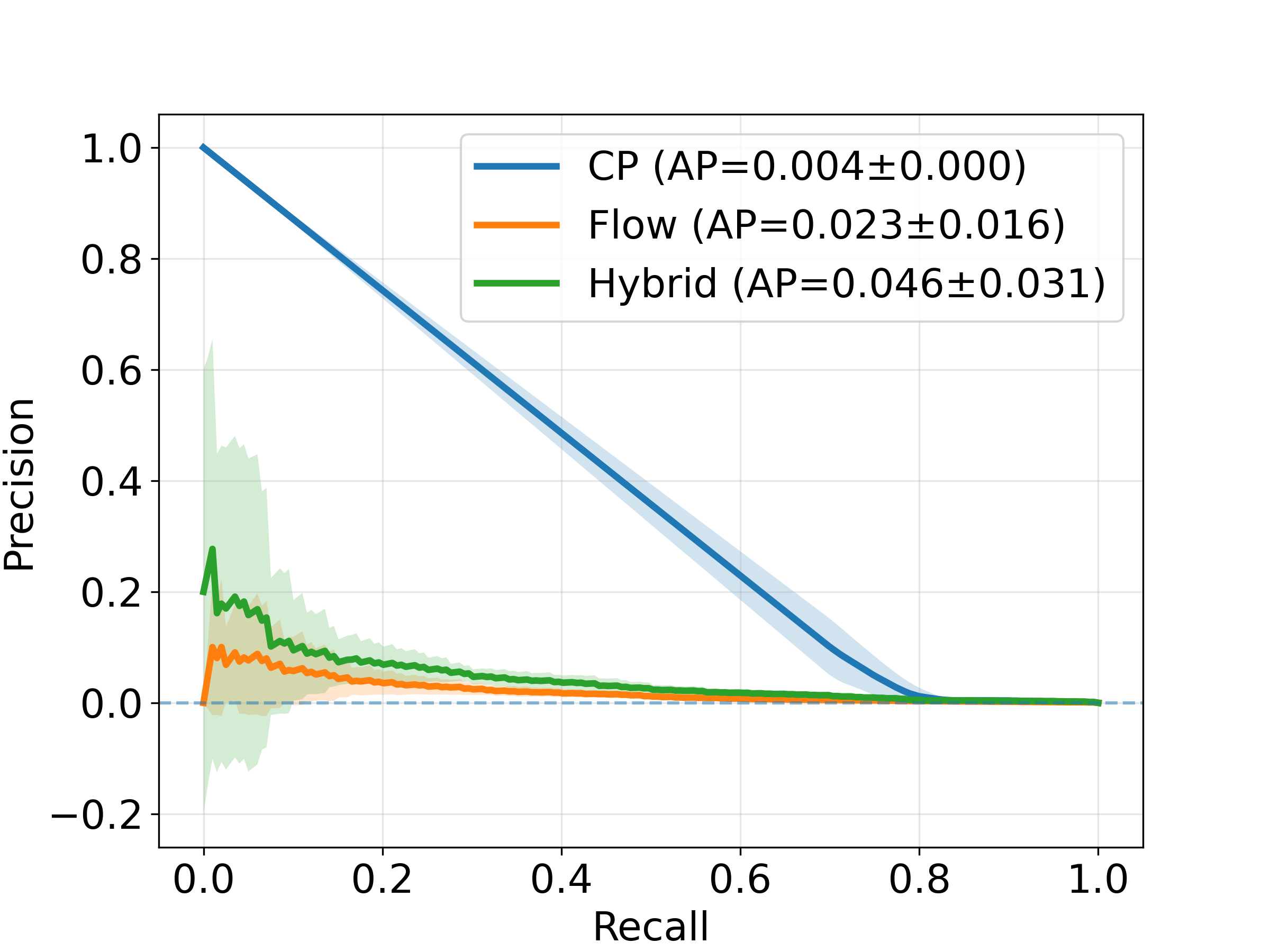}}
    \annotatedFigureText{0.4985,0.89}{black}{0.0523}{\large b} 
    \end{annotatedFigure} 
    \begin{annotatedFigure}
    {\includegraphics[width=5.5cm]{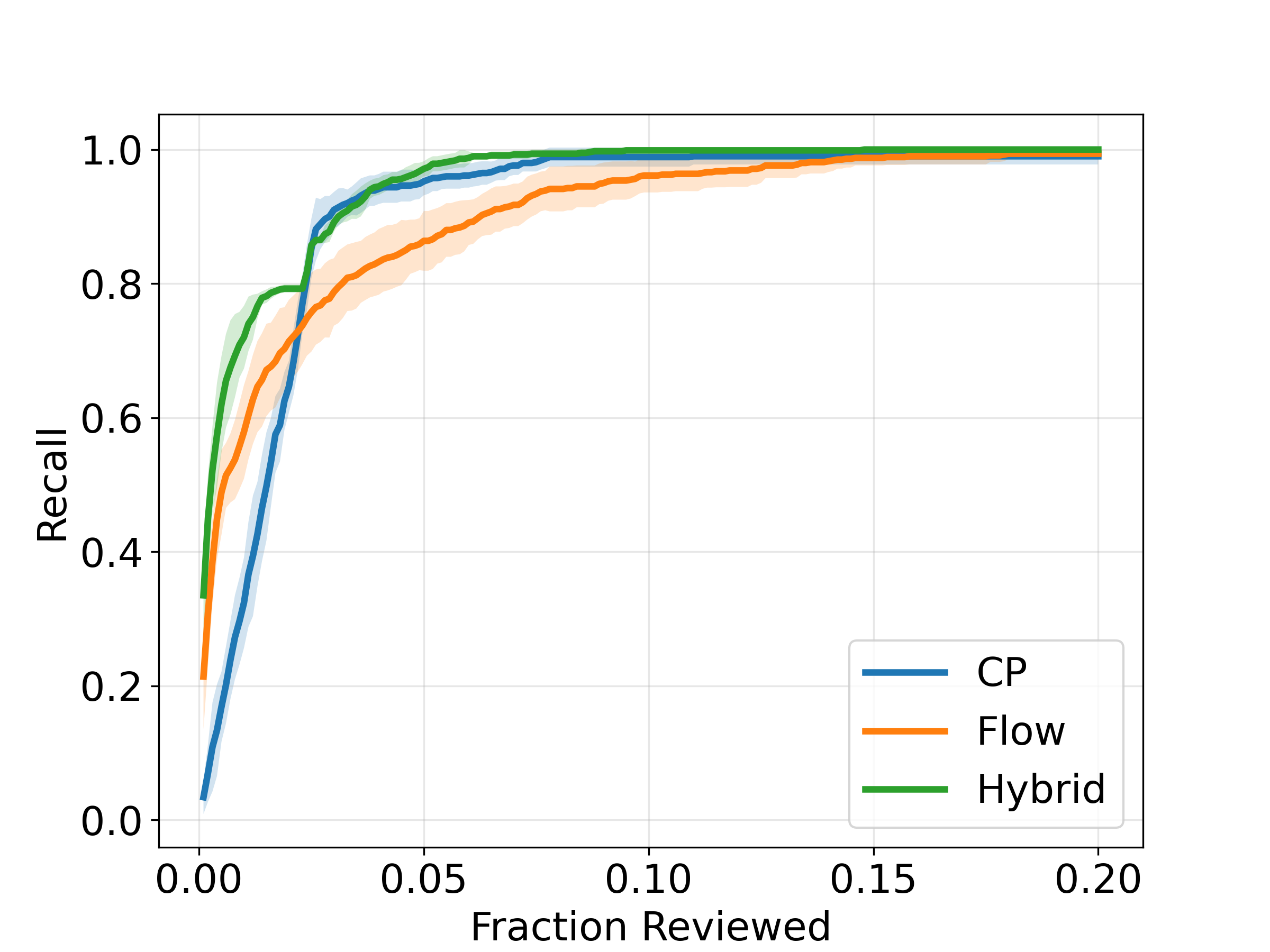}}
    \annotatedFigureText{0.4985,0.89}{black}{0.0523}{\large c} 
    \end{annotatedFigure} 
    \begin{annotatedFigure}
    {\includegraphics[width=5.5cm]{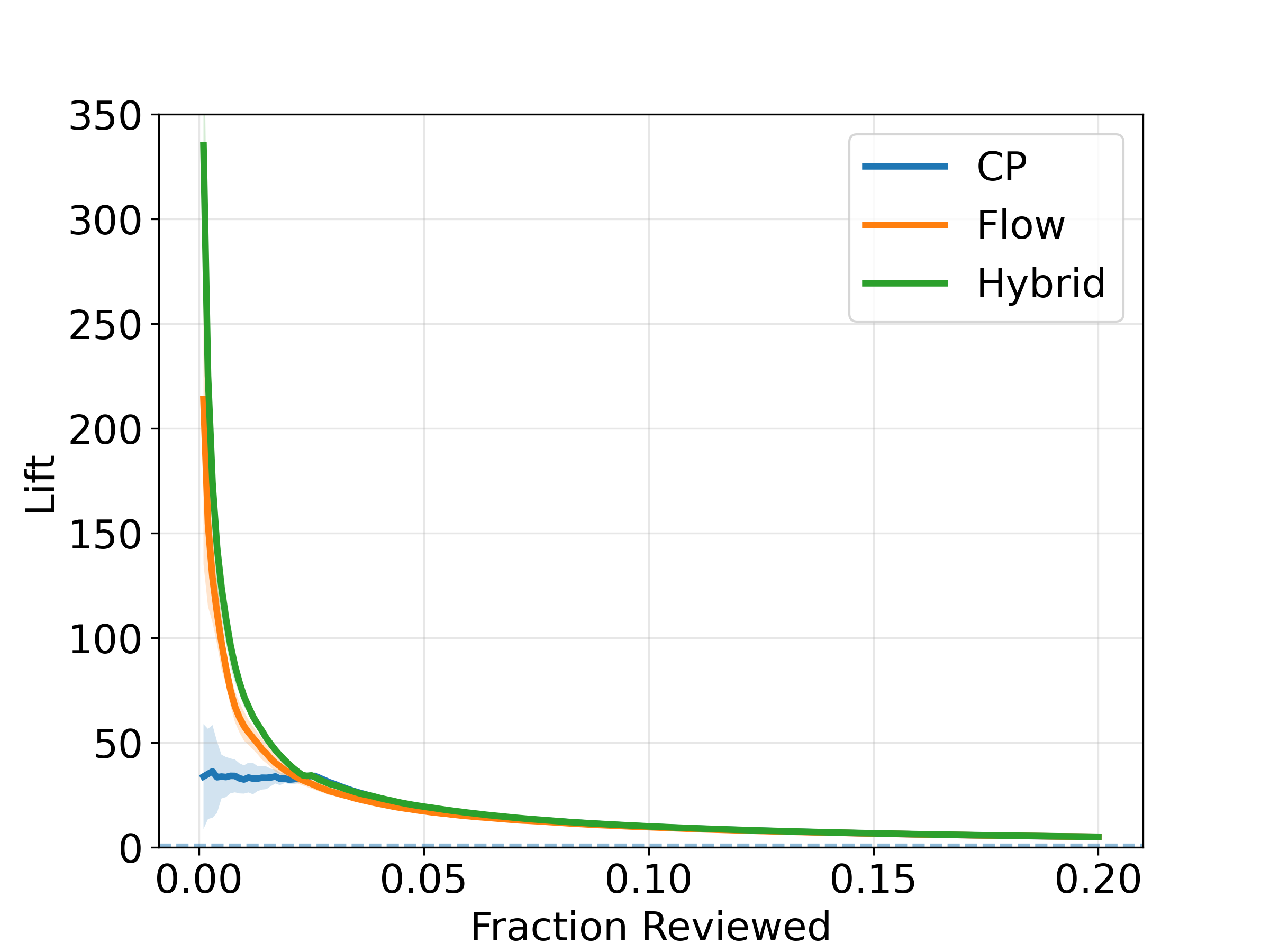}}
    \annotatedFigureText{0.4985,0.89}{black}{0.0523}{\large d} 
    \end{annotatedFigure} 
    \caption{Method comparisons using ensemble runs of the USDHs dataset for (a) ROC, (b) precision-recall, (c) recall versus fraction of dataset reviewed, and (d) lift versus fraction reviewed.}
    \label{USDHs}
\end{figure}

\begin{figure}
    \centering
    \begin{annotatedFigure}
    {\includegraphics[width=5.5cm]{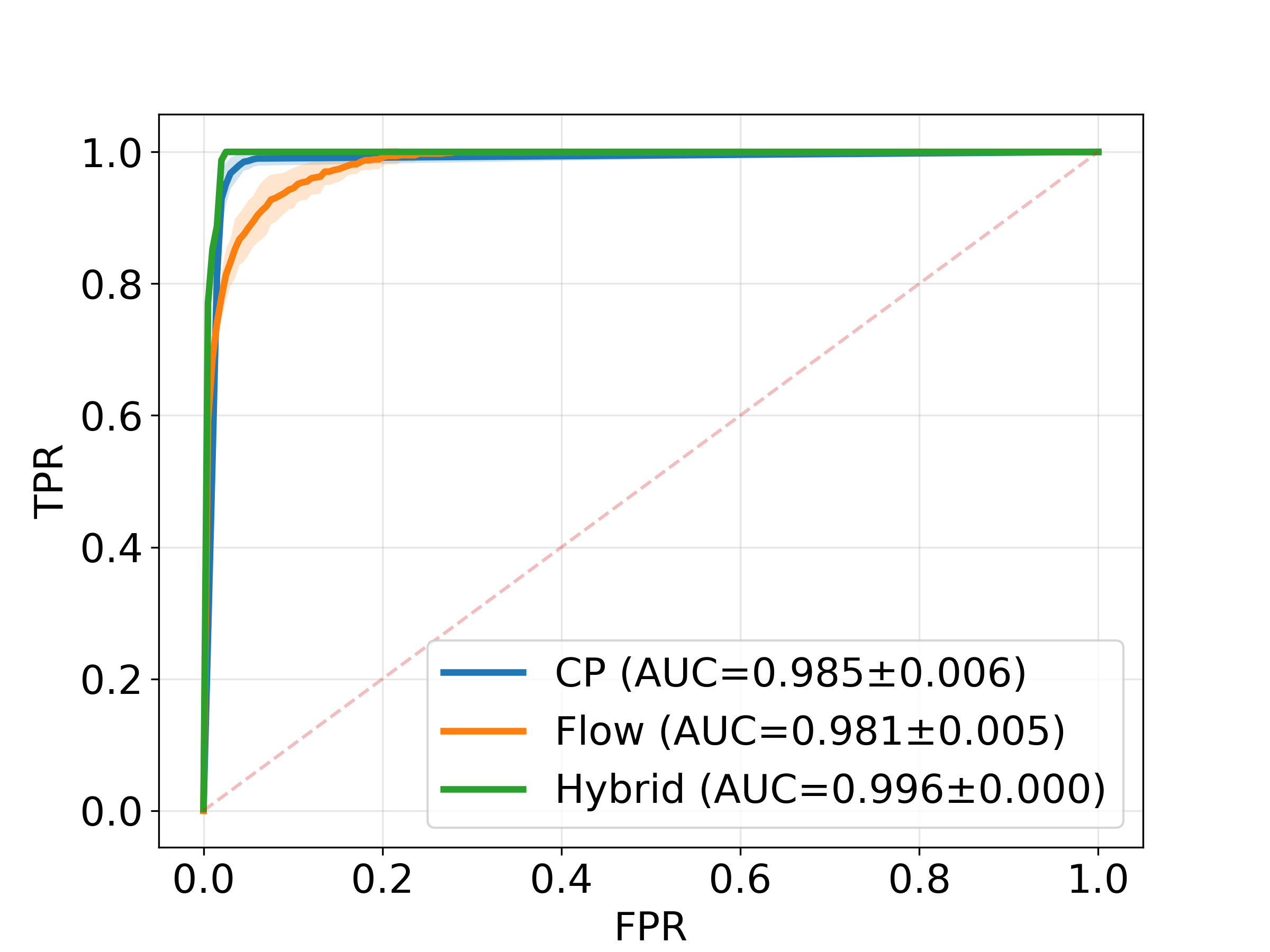}}
    \annotatedFigureText{0.4985,0.89}{black}{0.0523}{\large a} 
    \end{annotatedFigure} 
    \begin{annotatedFigure}
    {\includegraphics[width=5.5cm]{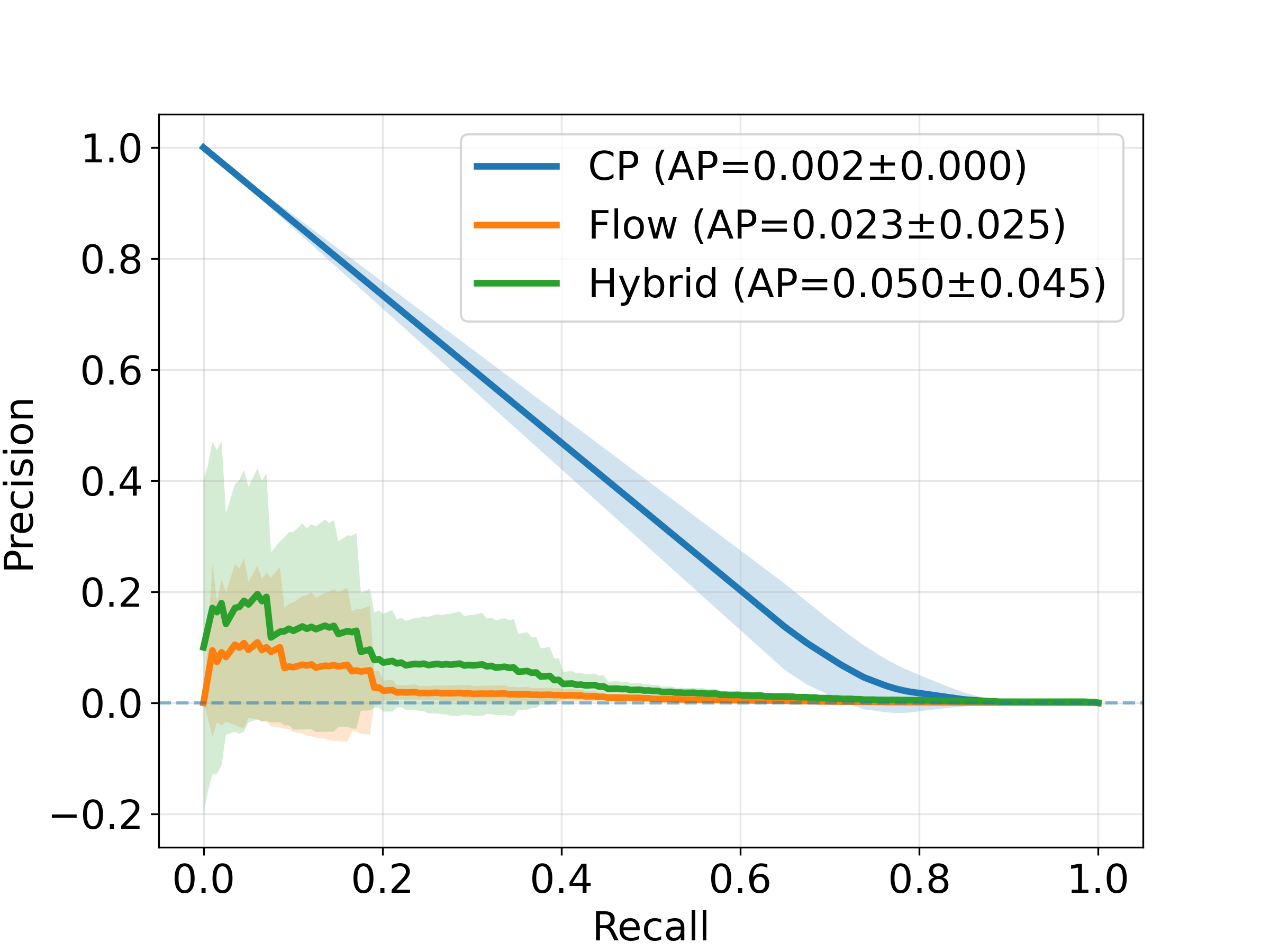}}
    \annotatedFigureText{0.4985,0.89}{black}{0.0523}{\large b} 
    \end{annotatedFigure} 
    \begin{annotatedFigure}
    {\includegraphics[width=5.5cm]{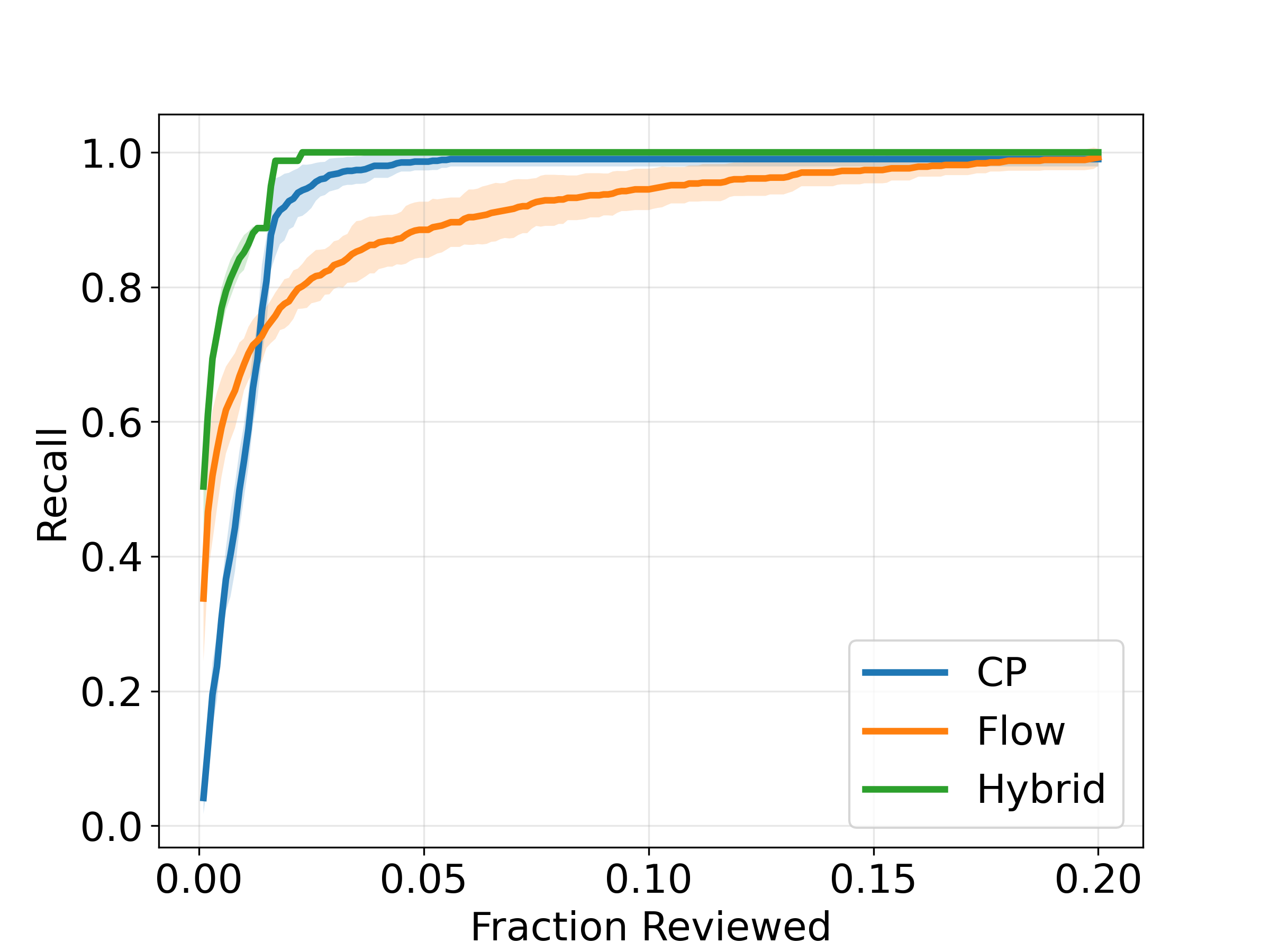}}
    \annotatedFigureText{0.4985,0.89}{black}{0.0523}{\large c} 
    \end{annotatedFigure} 
    \begin{annotatedFigure}
    {\includegraphics[width=5.5cm]{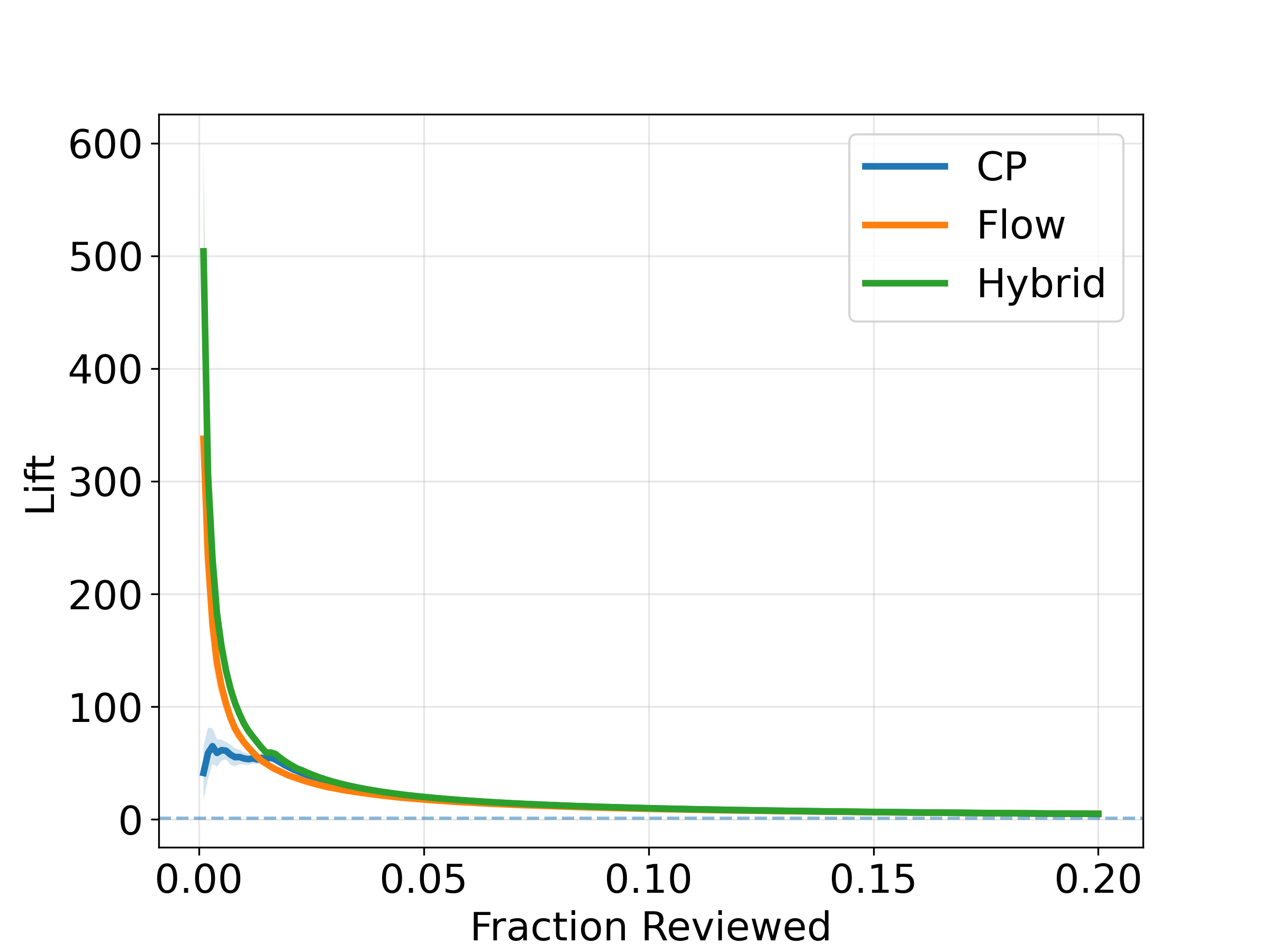}}
    \annotatedFigureText{0.4985,0.89}{black}{0.0523}{\large d} 
    \end{annotatedFigure} 
    \caption{Method comparisons using ensemble runs of the USDHDs dataset for (a) ROC, (b) precision-recall, (c) recall versus fraction of dataset reviewed, and (d) lift versus fraction reviewed.}
    \label{USDHDs}
\end{figure}

All results are reported over 10 ensemble runs for each tensor representation and each method. Quantities of interest are reported as mean $\pm 1\sigma$, corresponding to the variability across ensemble runs shown in Figures~\ref{US}--\ref{USDHDs}. We compare CP-APR, RealNVP, and the proposed HLSF framework using ROC-AUC, PR-AUC, recall as a function of the fraction of the dataset reviewed, and lift as a function of the fraction reviewed.
Table~\ref{table_method_comparison_auc} summarizes the ROC-AUC and PR-AUC values across all methods and tensor representations. ROC-AUC evaluates whether anomalous red-team events are ranked as more suspicious than benign events. For CP-APR, this corresponds to assigning lower Poisson survival-function p-values to anomalous events. For RealNVP, this corresponds to assigning larger negative log-likelihood scores to anomalous latent representations. HLSF combines both scores after normalization.

Overall, HLSF achieves the strongest ROC-AUC performance, obtaining the best ROC-AUC on all tensors except \textit{User--Destination} (UD). The improvement is especially clear for the higher-order tensors. On USD, USDs, USDHs, and USDHDs, HLSF consistently improves over both CP-APR and RealNVP, suggesting that the structural information captured by tensor factorization and the latent-density information captured by normalizing flows provide complementary anomaly signals. CP-APR remains competitive in ROC-AUC and performs best on UD, which is the least expressive representation considered in this study.

Precision-recall performance provides a stricter evaluation in this setting because the data are highly imbalanced. Although ROC-AUC values are often high, PR-AUC remains comparatively low because red-team events constitute a very small fraction of all observations. In operational cyber anomaly detection, PR-AUC is therefore especially important because it reflects the quality of the ranked alert list under severe class imbalance. RealNVP improves PR-AUC relative to CP-APR on most tensors, and HLSF achieves the best PR-AUC on USD, USDs, USDHs, and USDHDs. The only exception is UD, where CP-APR achieves the strongest ROC-AUC and PR-AUC.

Figures~\ref{US}--\ref{USDHDs} provide a more detailed comparison of the three methods across four metrics for each tensor representation. In each figure, panel (a) shows ROC performance, panel (b) shows precision-recall performance, panel (c) shows recall as a function of the fraction of the dataset reviewed, and panel (d) shows lift as a function of the fraction reviewed. Recall-versus-review and lift curves are operationally important because analysts typically investigate only a small fraction of events. We show the first 20\% of the reviewed data because the curves plateau after this range and performance at larger review fractions can be inferred. Higher early recall and lift indicate that a method identifies more red-team events while requiring fewer records to be reviewed.

Figure~\ref{US} shows results for the \textit{User--Source} (US) tensor. HLSF provides the highest ROC-AUC, while RealNVP achieves the highest PR-AUC by a small margin. Both RealNVP and HLSF substantially improve PR-AUC relative to CP-APR. In Figure~\ref{US}(b), CP-APR achieves high precision for a very small number of top-ranked anomalies, producing a visually tall curve near the earliest ranks. However, its precision deteriorates rapidly as recall increases, resulting in a lower overall average precision. By contrast, RealNVP and HLSF maintain stronger precision over a broader recall range. Figures~\ref{US}(c) and~\ref{US}(d) show that RealNVP and HLSF also provide stronger early recall and lift at small review fractions, indicating that they retrieve more anomalies when only a small portion of the dataset is inspected.

Figure~\ref{UD} shows results for the \textit{User--Destination} (UD) tensor. Unlike the other tensor representations, CP-APR outperforms RealNVP and HLSF in both ROC-AUC and PR-AUC. This result suggests that the UD representation does not provide enough contextual structure for the latent density model to improve over the tensor factorization baseline. The UD tensor preserves only user-destination relationships and omits source, status, and temporal context. As a result, the CP-APR latent representations are less informative for downstream density estimation, limiting the benefit of HLSF. Even so, Figures~\ref{UD}(c) and~\ref{UD}(d) show that RealNVP and HLSF remain competitive at very small review fractions, while CP-APR improves more substantially after approximately 8\% of the dataset has been reviewed.

Figure~\ref{USD} shows that adding source-destination context produces a clearer benefit for the hybrid approach. For the \textit{User--Source--Destination} (USD) tensor, HLSF achieves the highest ROC-AUC and PR-AUC and shows stronger early recall and lift than either standalone method. This indicates that once the tensor representation includes richer relational structure, the CP-APR factors provide useful latent event representations for RealNVP density estimation.

Figure~\ref{USDs} shows that including authentication status further improves detection performance. HLSF achieves the strongest ROC-AUC and PR-AUC, while RealNVP also substantially improves PR-AUC relative to CP-APR. This pattern suggests that status information provides useful behavioral context that is captured in the CP-APR latent factors and further exploited by the flow-based density model.

Figure~\ref{USDHs} shows results for the tensor that incorporates hour-of-day information. HLSF again achieves the best ROC-AUC and PR-AUC and maintains stronger early retrieval performance in the recall and lift curves. Adding temporal context helps distinguish routine authentication patterns from unusual behavior, and the hybrid model appears to benefit from this additional structure.

Figure~\ref{USDHDs} shows results for the highest-order tensor, which includes user, source, destination, hour, day, and status. HLSF achieves the strongest overall performance, with the highest ROC-AUC and PR-AUC among all methods. This tensor provides the most detailed behavioral representation, and the results indicate that the proposed hybrid model is especially effective when the tensor contains enough relational and temporal context for both low-rank structure and latent density information to be informative.

Taken together, the results show three main patterns. First, higher-order tensor representations generally improve anomaly detection because they encode richer behavioral context than lower-order matrices. Second, RealNVP often improves PR-AUC relative to CP-APR, indicating that density estimation over CP-derived latent representations can improve early anomaly ranking under severe class imbalance. Third, HLSF provides the most consistent overall performance, particularly for USD, USDs, USDHs, and USDHDs. This suggests that Poisson low-rank structure and latent-space density capture complementary anomaly signals: events that appear structurally plausible under CP-APR may still occupy low-density regions of latent space, while events with unusual relational structure remain detectable through the tensor model. By combining both views, HLSF yields a more robust ranking of anomalous events than either constituent method alone.

\section{Conclusions}

This work summarizes a novel method for improving anomaly detection in authentication datasets for cybersecurity applications. We demonstrate that when compared to both CP-APR and the RealNVP architecture of normalizing flows alone, the Hybrid Latent-Structural Fusion method (HLSF) outperforms the ability to detect anomalies when looking at the receiver operating characteristic (ROC), average precision scores (AP/PR), and recall and lift budgets as tensor dimensions increase. \cite{eren2023general} had previously shown that CP-APR outperformed other state-of-the-art supervised and unsupervised methods for anomaly detection. We validate their findings that higher-order representations enhance the detections of anomalies because tensor factorization techniques extract more predictive information over multiple dimensions \citep{eren2023general}. 

Specifically, the performance gains observed for USD, USDs, USDHs, and USDHDs are consistent with the proposed HLSF framework. As additional tensor modes are incorporated, the latent representations produced by CP-APR become increasingly informative, providing richer behavioral embeddings for density estimation by the normalizing flow. The growing advantage of HLSF relative to either constituent method suggests that structural and latent space anomaly signals become increasingly complementary as relational context is added. Our results ultimately show the expressive power when combining the the latent representation of the tensor provided by normalizing flows with the structural representation provided by CP-APR. 

The primary limitation of HLSF is that its effectiveness depends on the quality of the underlying tensor factorization. In low-dimensional representations such as UD, the latent embeddings may not contain sufficient information for effective density estimation. Future work should investigate adaptive fusion strategies and richer tensor representations that incorporate additional contextual attributes. Further work will also evaluate the generalizability of our method to other cyber datasets, as well as other anomaly detection domains.

\section{Acknowledgments}
This manuscript has been approved for unlimited release and has been assigned LA-UR-26-25362. This research was supported by the Advanced Simulation and Computing program of Los Alamos National Laboratory (LANL). LANL is operated by Triad National Security, LLC, for the National Nuclear Security Administration of the U.S. Department of Energy (Contract No. 89233218CNA000001). 











\vskip 0.2in


\section*{Declaration of competing interest}
The authors declare that they have no known competing financial interests or personal relationships that could have appeared to influence the work reported in this paper.

\bibliographystyle{cas-model2-names.bst}

\bibliography{references.bib}

\end{document}